\documentclass{article}
\usepackage{etoolbox}       %
\usepackage{tabularx}       %

\usepackage{graphicx}  %

\usepackage[parfill]{parskip}
\usepackage{amsmath,amsthm}
\usepackage{mathtools}  %
\usepackage[dvipsnames]{xcolor}         %
\usepackage{float}  %

\newtoggle{arxiv}
\toggletrue{arxiv}

  \usepackage[parfill]{parskip}
  \usepackage[citestyle=authoryear-comp,sorting=nyt,maxbibnames=99,backend=biber]{biblatex}
  \newcommand{\citep}{\parencite}
  \newcommand{\citet}{\textcite}
  \newlength{\defbaselineskip}
  \RequirePackage[T1]{fontenc}
  \RequirePackage[tt=false, type1=true]{libertine}
  \RequirePackage[varqu]{zi4}
  \RequirePackage[libertine]{newtxmath}

\usepackage{bm}

\usepackage[inline]{enumitem}
\usepackage{booktabs}
\usepackage[dvipsnames]{xcolor}  %
\usepackage{subcaption}
\usepackage{multirow}
\usepackage{wrapfig}
\usepackage{algorithm,algorithmicx,algpseudocode}
\usepackage{nicematrix}

\usepackage[frozencache,cachedir=.]{minted} %

\usepackage[colorlinks=true,linkcolor=blue,citecolor=magenta,urlcolor=red]{hyperref}
\usepackage[capitalise,noabbrev]{cleveref}

\usepackage{import}

\usepackage{pifont}%

\newcommand{\R}{\mathbb{R}}
\newcommand{\dt}{\Delta}
\newcommand{\A}{\bm{A}}
\newcommand{\B}{\bm{B}}
\newcommand{\C}{\bm{C}}

\newcommand{\dA}{\overline{\bm{A}}}
\newcommand{\dB}{\overline{\bm{B}}}

\newcommand{\dtAB}{(\dt, \A, \B)}

\newcommand{\dAB}{(\dA, \dB)}

\newcommand{\SM}{\makebox[2em]{\textbf{SM}}}
\newcommand{\SE}{\makebox[2em]{\textbf{SE}}}
\newcommand{\AM}{\makebox[2em]{\textbf{AM}}}
\renewcommand{\AE}{\makebox[2em]{\textbf{AE}}}

\newcommand{\para}[1]{\iftoggle{arxiv}{\paragraph{#1}}{\textbf{#1}}}

  \title{OTCE: Hybrid SSM and Attention with Cross Domain Mixture of Experts to construct Observer-Thinker-Conceiver-Expresser}
  \usepackage{authblk}
  \author[$^\ast$]{Jingze Shi}
  \author[$^\ast$]{Ting Xie}
  \author[$^\ast$]{Bingheng Wu}
  \author[$^\ast$]{Chunjun Zheng}
  \author[$^\ast$]{Kai Wang}
  \affil[$^\ast$]{Dalian Neusoft University of Information}
  \affil[ ]{{\texttt{losercheems@gmail.com}}}
  \date{}

\begin{document}

  \maketitle

  \begin{abstract}

\noindent
Recent research has shown that combining Mamba with Transformer architecture, which has selective state space and quadratic self-attention mechanism, outperforms using Mamba or Transformer architecture alone in language modeling tasks. The quadratic self-attention mechanism effectively alleviates the shortcomings of selective state space in handling long-term dependencies of any element in the sequence. We propose a position information injection method that connects the selective state space model with the quadratic attention, and integrates these two architectures with hybrid experts with cross-sharing domains, so that we can enjoy the advantages of both. We design a new architecture with a more biomimetic idea: Observer-Thinker-Conceiver-Expresser (OTCE), which can compete with well-known medium-scale open-source language models on a small scale in language modeling tasks.
\end{abstract}
  \section{Introduction}
\label{sec:introduction}

The Transformers (Attention is All You Need~\citep{transformer2017}) architecture is popular in modern deep learning language modeling, which can directly capture the relationship between any two elements in a sequence, effectively handle long-distance dependencies, however, the architecture has two main drawbacks. First, when processing long sequences, its self-attention mechanism's quadratic complexity and cache size limit the ability to handle long contexts. Second, Transformer lacks a single summary state, which means that each generated token must compute over the entire context.

Meanwhile, the Selective State Model (Mamba~\citep{gu2023mamba}) has emerged. Mamba achieves linear scaling of sequence length during training and maintains a constant state size during generation through its selective state update mechanism. Moreover, due to its linear recursive state update mechanism, Mamba has a single summary state. However, Mamba also has a major drawback, that is, its positional information depends on the implicit local positional information provided by the causal convolution, while long-distance dependencies depend on the matrix D that skips the connection between input and output. This makes Mamba perform poorly in capturing long-distance dependencies, such as correctly capturing input-output formats in context learning (ICL).

An efficient model must have a small state, and an effective model must have a state that contains all the necessary information from the context. To build a model that is both efficient and effective, the key is to design a state that is both compact and comprehensive in capturing the necessary context information. Our main goal is to combine self-attention and the Selective State Model to overcome their respective limitations, further combine them with a mixed expert with extensive general and cross-domain knowledge to build a better basic model architecture than Transformers or Mamba. The model has the ability to learn long context dependencies, aggregate states, and efficient reasoning. This paper proposes a bionic perspective, aiming to explore new model architectures by cleverly combining the Selective State Model with self-attention mechanism. This approach can fully utilize the advantages of the two mechanisms and promote language modeling in a more efficient and effective direction.

\para{Positional Information.} We first identified a key challenge in combining the Selective State Model with self-attention: the effective integration of positional information. In Mamba, positional information is provided by the implicit local positional information from the causal convolution, while self-attention itself cannot provide positional information, it relies on positional encoding to provide global positional context. To address this issue, we designed a relative positional information injection method that connects the inner product state of the Selective State Space with the inner product state of self-attention, allowing the input gate-state-output gate of the Selective State Space to make filtered relevant information have discrete relative positional information, and in self-attention, the discrete relative positional information is re-continuous to build long-term dependency relationships of relevant information. This method not only enables our model to have the ability to selectively process input sequences but also effectively capture long-distance dependency relationships. In complex multi-query associative recall tasks, our model trained on the same dataset outperforms larger-scale Mamba, Transformer models, and models that mix Mamba and Transformer without our proposed relative positional information injection method.

\para{Cross-Domain Mixture of Experts.} In human society, knowledge is widely distributed across different domains, and these domains are interconnected through common foundational knowledge and cross-domain connections. To simulate this phenomenon in the model, we designed two types of cross-domain mixed experts: Cohesive Cross-Domain Expert and Expansive Cross-Domain Expert. These experts (multi-layer perceptrons) store and transfer common foundational knowledge and cross-domain knowledge between different domains by sharing parameters. The Cohesive Cross-Domain Expert achieves close integration between domains by sharing linear layer parameters within all experts, which is more suitable for small-scale models with fewer experts because of its faster computation speed. The Expansive Cross-Domain Expert shares a complete MLP parameter, adding a common domain knowledge gate in each expert to control the flow of common domain knowledge into private MLP parameters, which is more suitable for large-scale models with more experts because it allows for more flexible adjustment and utilization of common knowledge. Experimental results show that the performance of these two cross-domain mixed experts on the same dataset is better than shared expert isolated mixed experts, confirming the effectiveness of our design in promoting cross-domain knowledge transfer and improving model generalization.

\para{Architecture Design.} From a biological perspective, the relationship between input and output can be described by observing, thinking, conceiving, and expressing four stages. Inspired by this idea, we designed a new architecture with a bionic perspective: Observer-Thinker-Conceiver-Expresser (\textbf{OTCE}). The OTCE architecture mimics the natural process of information processing in biology, aiming to optimize information processing and transmission in a modular way. In the Observer module, we use the selective state space's selection ability to filter out irrelevant information in the sequence to retain relevant information. In the Thinker module, we use the self-attention's ability to capture dependencies between any two elements in a sequence of any length, regardless of their position in the sequence, to build long-term dependency relationships. In the Conceiver module, we use the linear recursive state update mechanism of the state space to build a single summary state information. In the Expresser module, we combine the context-aware state information produced by self-attention considering all elements with the summary state information to build a context-aware summary state. We also studied the combination of the Selective State Space and self-attention with ordinary multi-layer perceptrons and cross-domain mixed experts, and finally determined a model combination with the lowest perplexity at the same parameter scale.

We empirically validated OTCE on multiple tasks, including semantic similarity evaluation, long-short text classification, natural language inference, keyword recognition, different domain selection tasks, context learning, and multi-query associative recall tasks. These experiments demonstrate the effectiveness of the OTCE architecture in handling complex language tasks.

\begin{figure}[H]
  \centering
  \includegraphics[width=1.0\linewidth]{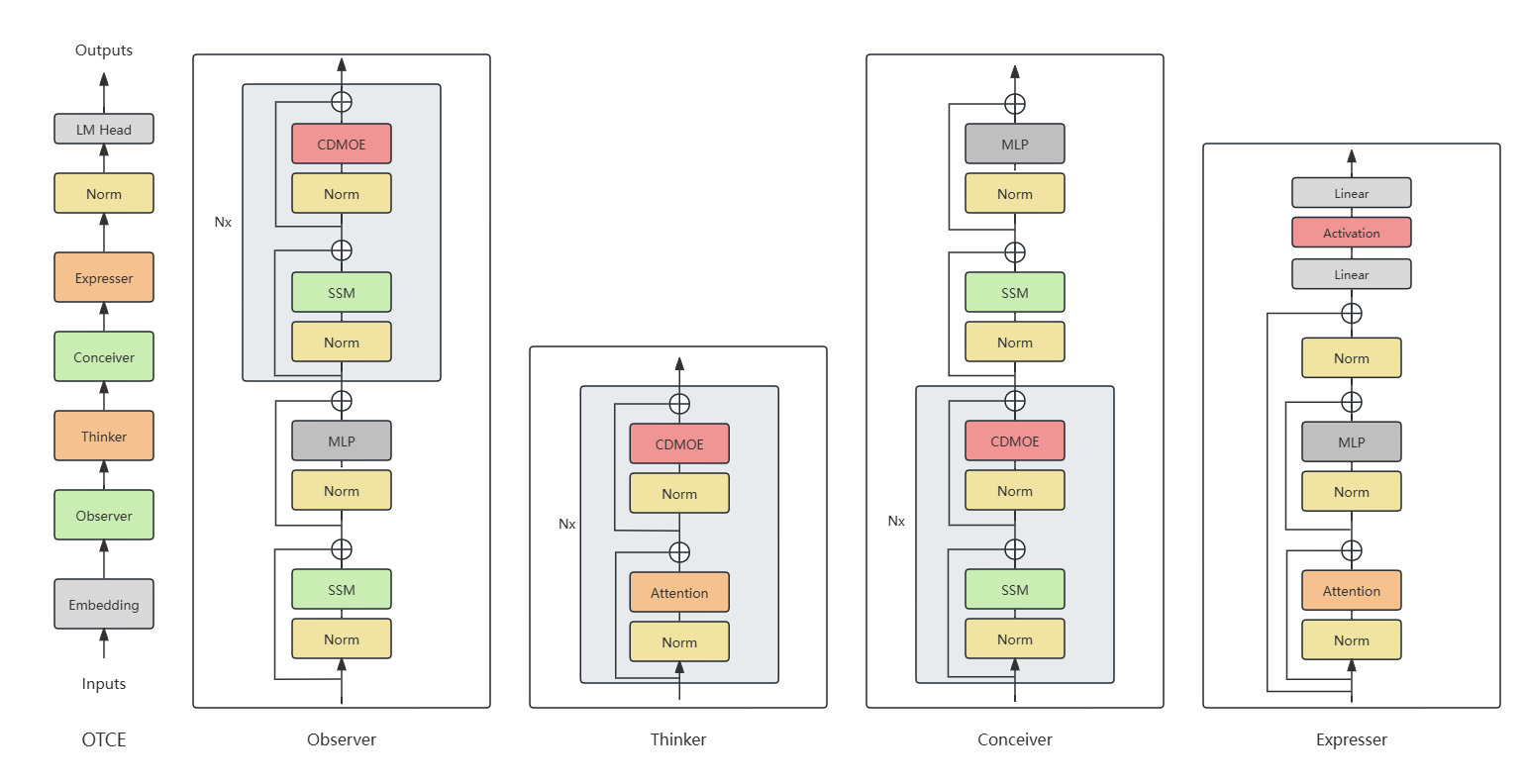}
  \caption{(\textbf{OTCE Architecture}.) OTCE demonstrates the overall combined architecture and process of using Observer, Thinker, Conceiver, and Expresser modules in language modeling tasks. Observer, Thinker, Conceiver, and Expresser show their internal combination of selective state space, self-attention, multi-layer perceptron, and cross-domain mixed experts.}
  \label{fig:otce}
\end{figure}
  \section{Background}
\label{sec:background}

\subsection{Selective State Space Models}
\label{sec:background:ssm}

Selective state space models~\citep{gu2023mamba} consider that a fundamental problem in sequence modeling is to compress the context into a smaller state. From this perspective, the attention mechanism of Transformers explicitly stores the entire context information, as if reviewing all previous inputs and generated tokens before writing each token. In contrast, RNNs only refer to a fixed number of previous tokens each time, which allows for faster writing but may forget key tokens.

In the precursor state space models of selective state space models, there are 4 parameters $(\dt, A, B, C)$, and they do not change with the input. These parameters control the following two stages:

\begin{minipage}[!ht]{.25\linewidth}
  \begin{subequations}
    \label{eq:ssm}
    \begin{align}
      h'(t) &= \A h(t) + \B x(t) \\
      y(t) &= \C h(t)
    \end{align}
  \end{subequations}
\end{minipage}
\begin{minipage}[H]{.28\linewidth}
  \begin{subequations}
    \label{eq:ssm:recurrence}
    \begin{align}
    \label{eq:ssm:recurrence:1}
      h_{t} &= \dA h_{t-1} + \dB x_t \\
    \label{eq:ssm:recurrence:2}
      y_t &= \C h_t
    \end{align}
  \end{subequations}
\end{minipage}
\begin{minipage}[H]{.40\linewidth}
  \begin{subequations}
    \label{eq:ssm:convolution}
    \begin{align}
      \label{eq:ssm:convolution:1}
      \bm{\overline{K}} &= (\bm{C}\bm{\overline{B}}, \bm{C}\bm{\overline{A}}\bm{\overline{B}}, \dots, \bm{C}\bm{\overline{A}}^{k}\bm{\overline{B}}, \dots) \\
      \label{eq:ssm:convolution:2}
      y &= x \ast \bm{\overline{K}}
    \end{align}
  \end{subequations}
\end{minipage}

\begin{itemize}[leftmargin=*,itemsep=0pt,topsep=0pt]
\item In the first stage (1a 1b), the continuous parameters $\dtAB$ are transformed into discrete parameters $\dAB$ through a fixed formula $\dA = f_A(\dt, \A)$ and $\dB = f_B(\dt, \A, \B)$, where $(f_A, f_B)$ is called the discretization rule. The most common is the zero-order hold (ZOH) rule, defined as $\dA = \exp(\dt \bm{A})$ and $\dB = (\dt \bm{A})^{-1} (\exp(\dt \bm{A}) - \bm{I}) \cdot \dt \bm{B}$.
\item In the second stage (2a 2b and 3a 3b), after the parameters are transformed from $(\dt, \A, \B, \C)$ to $(\dA, \dB, \C)$, linear recursion or global convolution can be used for computation.
\end{itemize}
State space models cannot update states selectively based on different input information.

The solution of selective state space models (Mamba~\citep{gu2023mamba}) is that, compared to state space models that compress all historical information, they design a simple selection mechanism that parameterizes the input of the state space model, making $(\dt, \B, \C)$ a function of the input.

\begin{figure*}[!ht]
  \begin{minipage}{.49\linewidth}
    \begin{algorithm}[H]
      \small
      \algrenewcommand\algorithmicrequire{\textbf{Input: }}
      \algrenewcommand\algorithmicensure{\textbf{Output: }}
      \caption{SSM (S4)}
      \label{alg:s4}
      \begin{algorithmic}[1]
        \Require $x : \mathtt{(B, L, D)}$
        \Ensure $y : \mathtt{(B, L, D)}$
        \State $\A : \mathtt{(D, N)} \gets \mathsf{Parameter}$

        \Comment{Represents structured $N \times N$ matrix}
        \State $\B : \mathtt{(D, N)} \gets \mathsf{Parameter}$
        \State $\C : \mathtt{(D, N)} \gets \mathsf{Parameter}$
        \State $\dt : \mathtt{(D)} \gets \tau_\dt(\mathsf{Parameter})$
        \State $\dA, \dB : \mathtt{(D, N)} \gets \mathsf{discretize}(\dt, \A, \B)$
        \State $y \gets \mathsf{SSM}(\dA, \dB, \C)(x)$

        \Comment{Time-invariant: recurrence or convolution}
        \State \textbf{return} $y$
      \end{algorithmic}
    \end{algorithm}
  \end{minipage}
  \begin{minipage}{.49\linewidth}
    \begin{algorithm}[H]
      \small
      \algrenewcommand\algorithmicrequire{\textbf{Input: }}
      \algrenewcommand\algorithmicensure{\textbf{Output: }}
      \caption{SSM + Selection (S6)}
      \label{alg:s6}
      \begin{algorithmic}[1]
        \Require $x : \mathtt{(B, L, D)}$
        \Ensure $y : \mathtt{(B, L, D)}$
        \State $\A : \mathtt{(D, N)} \gets \mathsf{Parameter}$

        \Comment{Represents structured $N \times N$ matrix}
        \State $\B : \textcolor{BrickRed}{\mathtt{(B, L, N)}} \gets \textcolor{BrickRed}{s_B(x)}$
        \State $\C : \textcolor{BrickRed}{\mathtt{(B, L, N)}} \gets \textcolor{BrickRed}{s_C(x)}$
        \State $\dt : \textcolor{BrickRed}{\mathtt{(B, L, D)}} \gets \tau_\dt(\mathsf{Parameter} \textcolor{BrickRed}{+ s_\dt(x)})$
        \State $\dA, \dB : \textcolor{BrickRed}{\mathtt{(B, L, D, N)}} \gets \mathsf{discretize}(\dt, \A, \B)$
        \State $y \gets \mathsf{SSM}(\dA, \dB, \C)(x)$

        \Comment{\textcolor{BrickRed}{Time-varying}: recurrence (\textcolor{BrickRed}{\emph{scan}}) only}
        \State \textbf{return} $y$
      \end{algorithmic}
    \end{algorithm}
  \end{minipage}
\end{figure*}

\begin{itemize}[leftmargin=*,itemsep=0pt,topsep=0pt]
\item The sizes of the matrices $\B$ that affect the input and $\C$ that affect the state change from $(D, N)$ to $(B, L, N)$, i.e., from the fixed input vector dimension $D$ and the state dimension $N$ of the SSM to the variable input batch size $B$, sequence length $L$, and state dimension $N$. The size of $\dt$ changes from $D$ to $(B, L, D)$, which means that each token in a batch has a unique $\dt$. The matrices $\B$, $\C$, and $\dt$ at each position are different, which means that each input token has different $\B$ and $\C$, solving the content-aware problem.
\item The dimension changes are specifically implemented by $s_B(x) = Linear_N(x)$, $s_C(x) = Linear_N(x)$, and $s_\dt(x) = Linear_D(x)$, which sequentially make $(\dt, B, C)$ data-dependent.
\item Although $A$ has not become data-dependent, after the discretization operation of the SSM, $(\dA, \dB)$ will be transformed into a data-dependent tensor of $(B, L, D, N)$ through vector products, which is a parameter-efficient way to achieve data dependency. That is, after the discretization of $A$, $\dA = \exp(\dt A)$, the data dependency of $\dt$ can make the overall $\dA$ related to the input.
\end{itemize}

This allows the model to selectively process information to focus on or ignore specific information and adaptively adjust its behavior.

\subsection{Attention and SSM}

Attention is a mechanism that computes the relevance scores between each element in a sequence and all other elements, allowing each element to "attend" to other elements. The most important variant of attention is softmax self-attention.

\begin{align}
Y &= \operatorname*{softmax}({QK^\top}) \cdot V
\end{align}

\begin{figure}[H]
  \centering
  \begin{minipage}{.49\linewidth}
    \begin{algorithm}[H]
      \small
      \algrenewcommand\algorithmicrequire{\textbf{Input: }}
      \algrenewcommand\algorithmicensure{\textbf{Output: }}
      \caption{Self-Attention}
      \label{alg:self_attention}
      \begin{algorithmic}
        \Require $x : \mathtt{(B, L, D)}$
        \Ensure $y : \mathtt{(B, L, D)}$
        \State $Q : \mathtt{(B, L, D)} \gets xW_Q$
        \State $K : \mathtt{(B, L, D)} \gets xW_K$
        \State $V : \mathtt{(B, L, D)} \gets xW_V$
        \State $A : \mathtt{(B, L, L)} \gets \operatorname*{softmax}({QK^\top})$
        \State $y : \mathtt{(B, L, D)} \gets AV$
        \State \textbf{return} $y$
      \end{algorithmic}
    \end{algorithm}
  \end{minipage}
\end{figure}

A notable feature of self-attention is that it can capture dependencies between any positions in the input sequence without being limited by distance, giving it an advantage in handling arbitrary element dependencies in long sequences.

Many variants of attention have been proposed, all of which are based on the attention score as the core. \textbf{Linear attention}~\citep{katharopoulos2020transformers} discards softmax by folding it into the kernel feature map and rewrites it using the kernel property of matrix multiplication as $(QK^\top) \cdot V = Q \cdot (K^\top V)$. In the case of causal (autoregressive) attention, they show that when the causal mask is merged on the left side as $(L \circ QK^\top) \cdot V$, where $L$ is a lower triangular matrix, the right side can be expanded into a recursive form. In \textbf{Transformers are SSMs}~\citep{mamba2}, the simple computation of scalar structured SSM is proven to be equivalent to quadratic masked attention by materializing the half-separable matrix $M = L \circ CB^\top = L \circ QK^\top$ and performing quadratic matrix-vector multiplication.

\begin{align}
\label{eq:attention_ssm}
Y = (L \circ QK^\top) \cdot V = (L \circ CB^\top) \cdot X
\end{align}

\subsection{Mixture of Experts}

The mixture of experts architecture~\citep{jiang2023mistral} aims to train a larger model with fewer training steps under limited computational resources, often yielding better results than training a smaller model with more steps. The mixture of experts architecture consists of two key components:
\begin{itemize}[leftmargin=*,itemsep=0pt,topsep=0pt]
\item The MOE layer, which replaces the feedforward network layer in traditional Transformer models. Each expert is an independent neural network, typically an MLP, but it can also be a more complex network structure or even the MOE layer itself.
\item The gating network or router, which determines which tokens are sent to which expert. A token can be sent to multiple experts. The router consists of learnable parameters and is trained together with other parts of the network.
\end{itemize}

To ensure that experts receive non-overlapping concentrated knowledge, DeepSeekMoE \citep{dai2024deepseekmoe} isolates K experts for sharing, aiming to capture common knowledge and reduce redundancy.

\begin{figure}[!ht]
  \centering
  \includegraphics[width=0.70\linewidth]{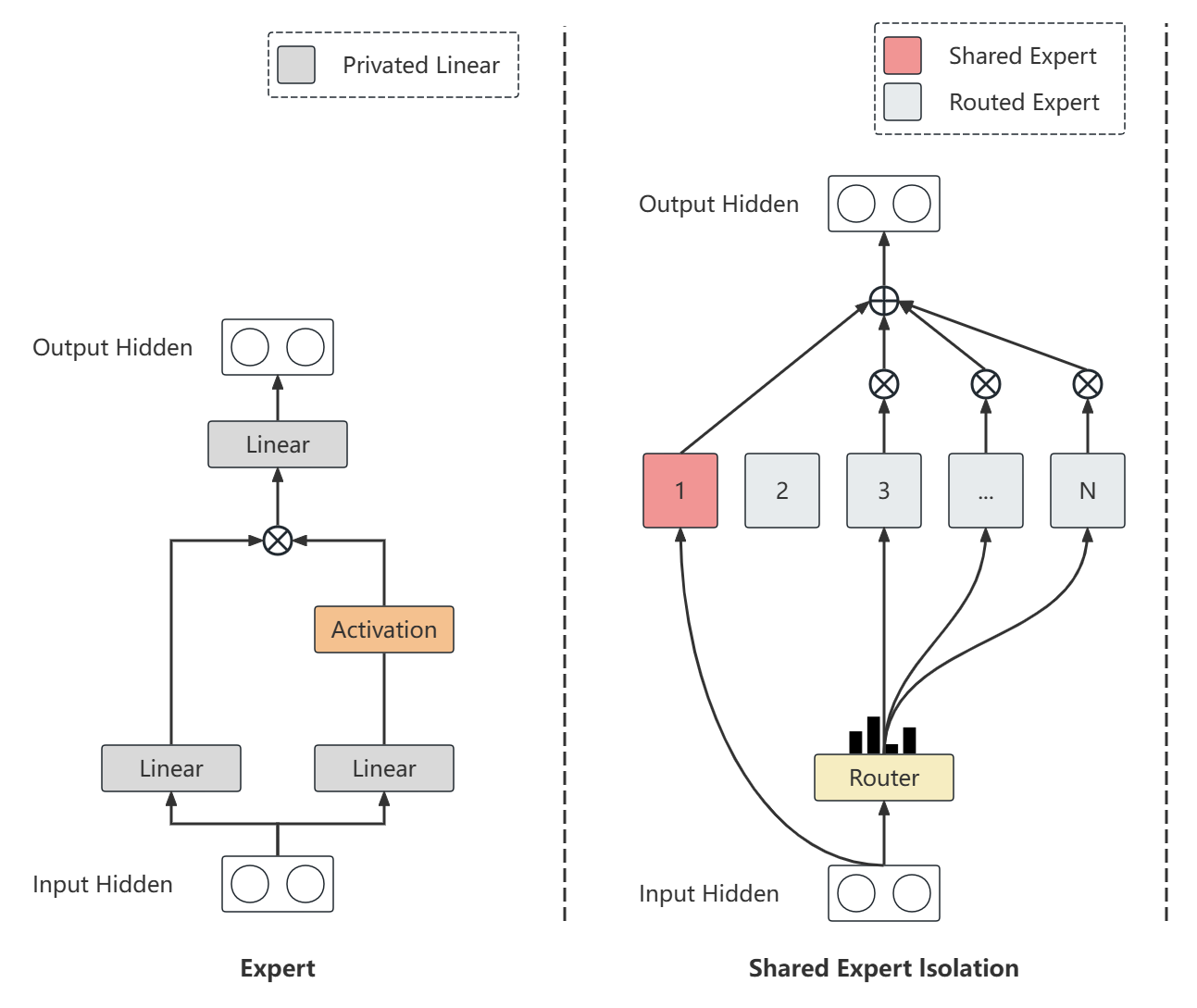}
  \caption{(\textbf{Shared Expert Isolation}.) The hidden state of the entire sequence calculated by the shared expert is added to the hidden state of each token calculated by the most relevant routing expert. Each token is composed of the shared expert state and the routing expert state.}
  \label{fig:shared_expert_isolation}
\end{figure}
\clearpage
  \section{Methods}

\subsection{Positional Information}
\label{sec:positional_information}

\paragraph{Rotational Positional Encoding.}
The so-called rotational positional encoding, in which the absolute positional embedding is removed, and the rotational positional encoding is added at each layer of the network~\citep{kexuefm-8265}, is a form of relative positional encoding implemented in the form of absolute positional encoding using the idea of complex numbers.

\begin{subequations}
    \label{eq:attention_rope}
    \begin{align}
        Q_{m} &= f_{Q} (x_{m}, m) \\
        K_{n} &= f_{K} (x_{n}, n)
    \end{align}
\end{subequations}

$Q_{m}$ represents the query vector integrated with the positional information $m$ of the $m$-th token corresponding to the word embedding $x_{m}$, and $K_{n}$ represents the key vector integrated with the positional information $n$ of the $n$-th token corresponding to the word embedding $x_{n}$.

\paragraph{Rotational Positional Encoding for SSM.}
In the matrix transformation form of the state space model mentioned in \textbf{Transformers are SSMs}~\citep{mamba2},
by definition, $h_0 = B_0 x_0$.
By induction,

\begin{align*}
    h_t &= A_t \dots A_1 B_0 x_0 + A_t \dots A_2 B_1 x_1 + \dots + A_t A_{t-1} B_{t-2} x_{t-2} + A_t B_{t-1} x_{t-1} + B_t x_t
    \\&= \sum_{s=0}^t A_{t:s}^\times B_s x_s
\end{align*}

Multiplying by $C_t$ to produce $y_t$, and vectorizing the equation to $t \in [\mathtt{T}]$ ($\mathtt{T}$ is the sequence length), we derive the matrix transformation form of SSM.

\begin{equation}
\label{eq:ssm-matrix}
\begin{aligned}
    y_t &= \sum_{s=0}^t C_t^{\top} A_{t:s}^\times B_s x_s
    \\
    y &= \mathsf{SSM}(A, B, C)(x) = Mx
    \\
    M_{ji} &\coloneqq C_j^{\top} A_{j} \cdots A_{i+1} B_{i}
\end{aligned}
\end{equation}

Then the matrix form of SSM is represented using SSS (Sequentially Semiseparable) as $M = \mathsf{SSS}(A, B, C)$, where $M_{ji} = C_j^{\top} A_{j:i} B_i$~\ref{eq:ssm-matrix}, and then considering $A$ is just a scalar, rearranged as

\begin{align*}
    M_{ji} = A_{j:i} \cdot (C_j^{\top}B_i)
\end{align*}

Vectorized as

\begin{equation}
\label{eq:ssm-matrix-vectorized}
    \begin{aligned}
        L &\coloneqq \mathsf{1SS}(a) \\
        M &= L \circ (C B^{\top}) \\
    \end{aligned}
\end{equation}

Finally, it is proved that the matrix transformation form of SSM is equivalent to Attention $(L \circ QK^\top) \cdot V = (L \circ CB^\top) \cdot X$~\ref{eq:attention_ssm}.

Now we have enough theoretical support to give rotational positional encoding to the $C$ and $B$ matrices in SSM.

\begin{subequations}
    \label{eq:ssm_rope}
    \begin{align}
        C_{j} &= f_{C} (x_{j}, j) \\
        B_{i} &= f_{B} (x_{i}, i)
    \end{align}
\end{subequations}

$C_{j}$ represents the output weight matrix integrated with the positional information $j$ of the $j$-th token corresponding to the word embedding $x_{j}$, and $B_{i}$ represents the input weight matrix integrated with the positional information $i$ of the $i$-th token corresponding to the word embedding $x_{i}$.

To utilize the relative positional information between tokens, we assume that the inner product operation between $C_{j}$ and $B_{i}$ vectors can be represented by a function $g$, the input of this function $g$ is the word embedding vectors $x_{j}$ and $x_{i}$, and their relative positional information $j - i$, the relationship between $C_{j}$ and $B_{i}$ inner product and their relative positional information $j - i$ is defined as

\begin{equation}
    \label{eq:ssm_rope_g}
    <f_{C}(x_{j}, j), f_{B}(x_{i}, i)> = g(x_{j}, x_{i}, j - i)
\end{equation}

We can use the geometric properties of vectors on a 2-dimensional plane, combined with the properties of complex numbers, to prove

\begin{align}
    \label{eq:ssm_rope_g_final}
    <f_{C}(x_{j}, j), f_{B}(x_{i}, i)> &= 
        \begin{bmatrix} 
            C_{j}^{(1)} & C_{j}^{(2)}
        \end{bmatrix}
        \begin{bmatrix} 
            \cos((j - i) \theta) & -\sin((j - i) \theta) \\
            \sin((j - i) \theta) & \cos((j - i) \theta)
        \end{bmatrix}
        \begin{bmatrix} 
            B_{i}^{(1)} \\
            B_{i}^{(2)}
        \end{bmatrix}
\end{align}

The proof process is detailed in Appendix~\ref{sec:rope_for_ssm}.

\begin{figure}[!ht]
    \centering
    \includegraphics[width=0.8\textwidth]{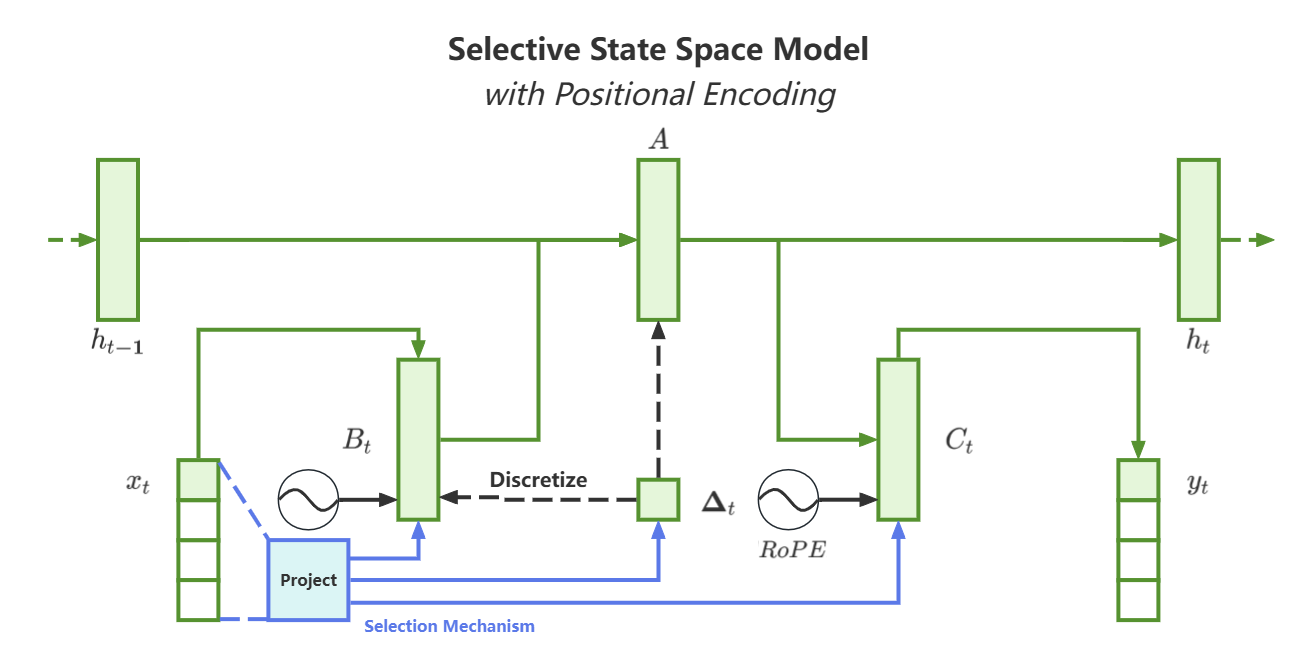}
    \caption{
        (\textbf{RoPE for SSM}.) 
        The structure diagram of applying rotational positional encoding to the Selective State Space Model.
    }
    \label{fig:selective_state_space_model_with_positional_encoding}
\end{figure}

In this way, we can apply rotational positional encoding to SSM as shown in Figure~\ref{fig:selective_state_space_model_with_positional_encoding} and Algorithm~\ref{alg:s6_with_rope}.

Recalling the algorithm of the Selective State Space Model~\ref{alg:s6}, the matrix $B$ controls whether to let the input $x_t$ affect the state $h_t$, after adding rotational positional encoding to the matrix $B$, the matrix $B$ selectively filters the input $x_t$ to make $h_t$ have discrete positional information, the matrix $A$ controls whether the state $h_{t-1}$ affects the state $h_t$, that is, the matrix $A$ also allows the position information of important states stored in the past state to affect the current state, the matrix $C$ controls whether the state $h_t$ affects the output $y_t$, after the inner product operation between the matrix $C$ and the $h_t$ with discrete positional information, the output $y_t$ has discrete relative positional information, which is undoubtedly an important supplement to the Selective State Space Model, a recursive model that is difficult to perform comprehensive context association and memory tasks (including state interference, no global position as a reference factor in sequence transformation, etc.).

\begin{figure}[H]
    \centering
    \begin{minipage}{.49\linewidth}
    \begin{algorithm}[H]
      \small
      \algrenewcommand\algorithmicrequire{\textbf{Input: }}
      \algrenewcommand\algorithmicensure{\textbf{Output: }}
      \caption{S6 with RoPE}
      \label{alg:s6_with_rope}
      \begin{algorithmic}
      \Require $x : \mathtt{(B, L, D)}, pos_{id} : \mathtt{(B, L)}$
      \Ensure $y : \mathtt{(B, L, D)}$
      \State $\A : \mathtt{(D, N)} \gets \mathsf{Parameter}$
      
      \Comment{Represents structured $N \times N$ matrix}
      \State $cos, sin : \textcolor{BrickRed}{\mathtt{(L, N)}} \gets \textcolor{BrickRed}{Ro_{emb}(\mathtt{N})}$
      \State $\B : \textcolor{BrickRed}{\mathtt{(B, L, N)}} \gets \textcolor{BrickRed}{RoPE(cos, sin, pos_{id}, s_B(x))}$
      \State $\C : \textcolor{BrickRed}{\mathtt{(B, L, N)}} \gets \textcolor{BrickRed}{RoPE(cos, sin, pos_{id}, s_C(x))}$
      
      \State $\dt : {\mathtt{(B, L, D)}} \gets \tau_\dt(\mathsf{Parameter} {+ s_\dt(x)})$
      \State $\dA, \dB : {\mathtt{(B, L, D, N)}} \gets \mathsf{discretize}(\dt, \A, \B)$
      \State $y \gets \mathsf{SSM}(\dA, \dB, \C)(x)$

      \Comment{{Time-varying}: recurrence ({\emph{scan}}) only}
      \State \textbf{return} $y$
      \end{algorithmic}
    \end{algorithm}
    \end{minipage}
\end{figure}

\begin{figure}[H]
    \centering
    \begin{minipage}{0.45\textwidth}
        \centering
        \includegraphics[width=1.0\textwidth]{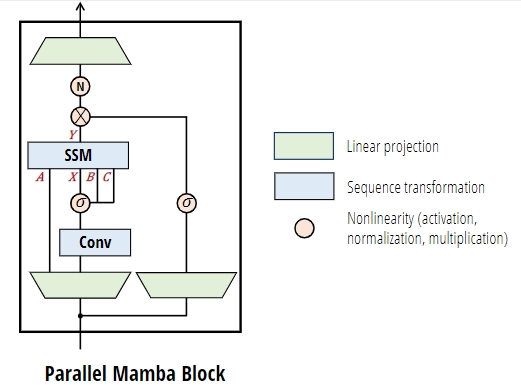}
    \end{minipage} \hfill
    \begin{minipage}{0.45\textwidth}
        \centering
        \includegraphics[width=1.0\textwidth]{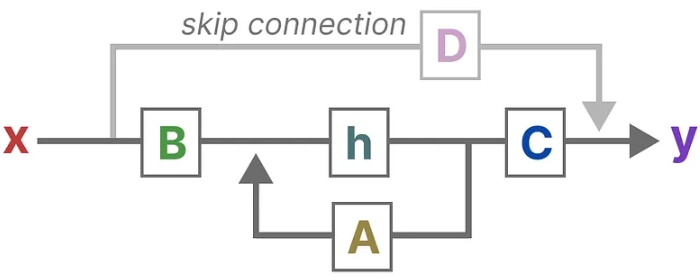}
    \end{minipage}
    \caption{
            (\textbf{Mamba's Positional Information}.) 
            Mamba provides continuous relative positional information for the matrix $D$ that skips the connection between the input and output of SSM by convolution.
        }
    \label{fig:mamba_and_ssm}
\end{figure}

\paragraph{Converting Discrete Position Information to Continuous Position Information for Attention Weighting.}
Although the $y$ output by this kind of selective state space model with positional encoding is all effective information, the positional information is discrete.
If we don't provide continuous relative positional information like Mamba~\citep{gu2023mamba} as shown in Figure~\ref{fig:mamba_and_ssm}, by convolution to provide continuous relative positional information for the matrix $D$ that skips the connection between the input and output of SSM before the SSM matrix transformation,
we want to directly take this $y$ with only discrete relative positional information for self-attention weighting to build long-term dependency relationships is completely insufficient.

\begin{figure}[H]
    \centering
    \includegraphics[width=0.8\textwidth]{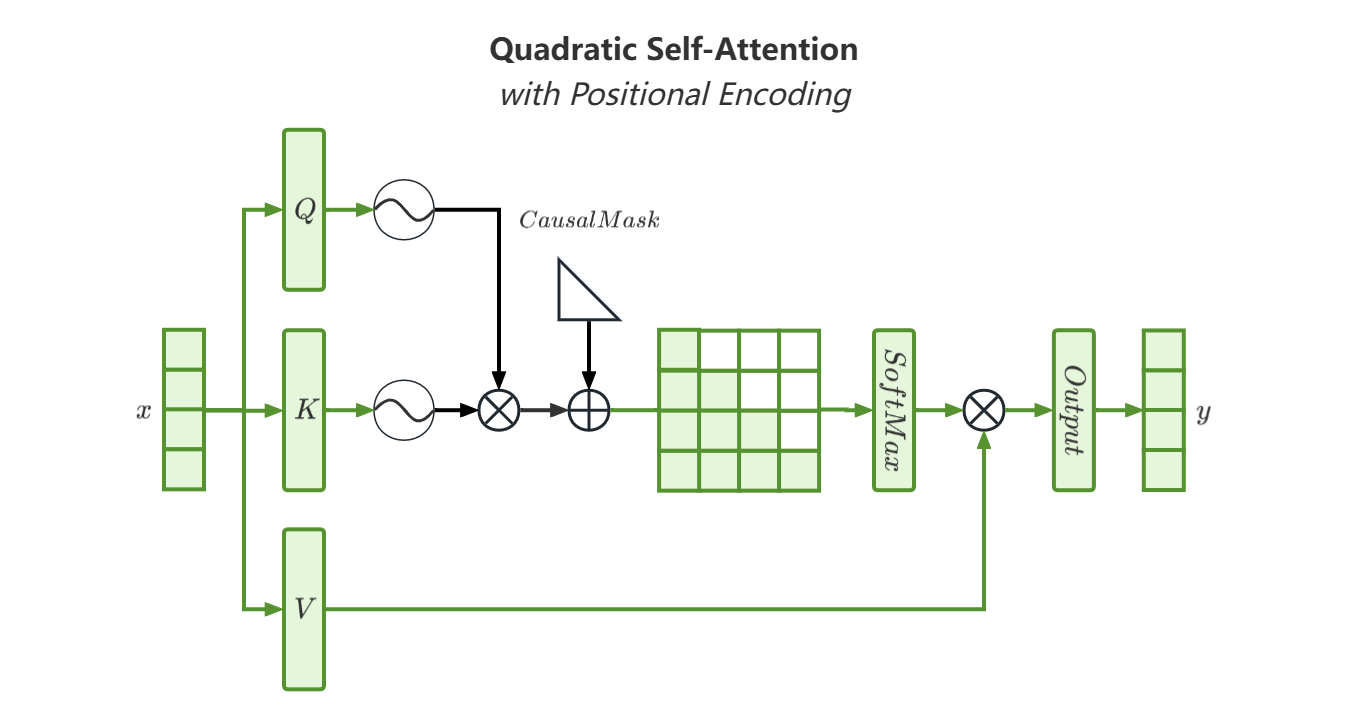}
    \caption{
        (\textbf{RoPE for Quadratic Self-Attention}.)
        Applying rotational positional encoding to the quadratic self-attention mechanism, reintroducing continuous positional encoding information before the $QK$ inner product operation.
    }
    \label{fig:quadratic_self_attention_with_positional_encoding}
\end{figure}

If a position has invalid positional information,
after the inner product operation between $Q$ and $K$,
this kind of vacancy interference will be amplified.
Therefore, if we want to remove convolution operations or even remove the matrix $D$,
we need to reintroduce the continuous positional encoding information as shown in Equation~\ref{eq:attention_rope_g_final} before performing the quadratic self-attention inner product operation again on $y$ that has discrete relative positional information, as shown in Figure~\ref{fig:quadratic_self_attention_with_positional_encoding}.

\begin{align}
    \label{eq:attention_rope_g_final}
    <f_{Q}(x_{m}, m), f_{K}(x_{n}, n)> &= 
        \begin{bmatrix} 
            Q_{m}^{(1)} & Q_{m}^{(2)}
        \end{bmatrix}
        \begin{bmatrix} 
            \cos((m - n) \theta) & -\sin((m - n) \theta) \\
            \sin((m - n) \theta) & \cos((m - n) \theta)
        \end{bmatrix}
        \begin{bmatrix} 
            K_{n}^{(1)} \\
            K_{n}^{(2)}
        \end{bmatrix}
\end{align}

\subsection{Cross Domain Mixture of Experts}
\paragraph{Structure of a Single Expert.}
GLU~\citep{glu2020} tried gated linear units and variants in the feedforward network layer of the Transformer, and found that some variants are more effective than common activation functions such as ReLU and GELU. It consists of the product of two linear parameters, one of which is passed through an activation function. Taking the $Swish$ function as an example, it is defined as $SwiGLU(x, W, V, b, c) = Swish(xW + b) \cdot (xV + c)$, and its feedforward network layer can be defined as $FFN_{SwiGLU}(x, W, V, W_2) = (Swish(xW) \cdot (xV))W_2$.

\begin{figure}[H]
    \centering
    \includegraphics[width=0.9\textwidth]{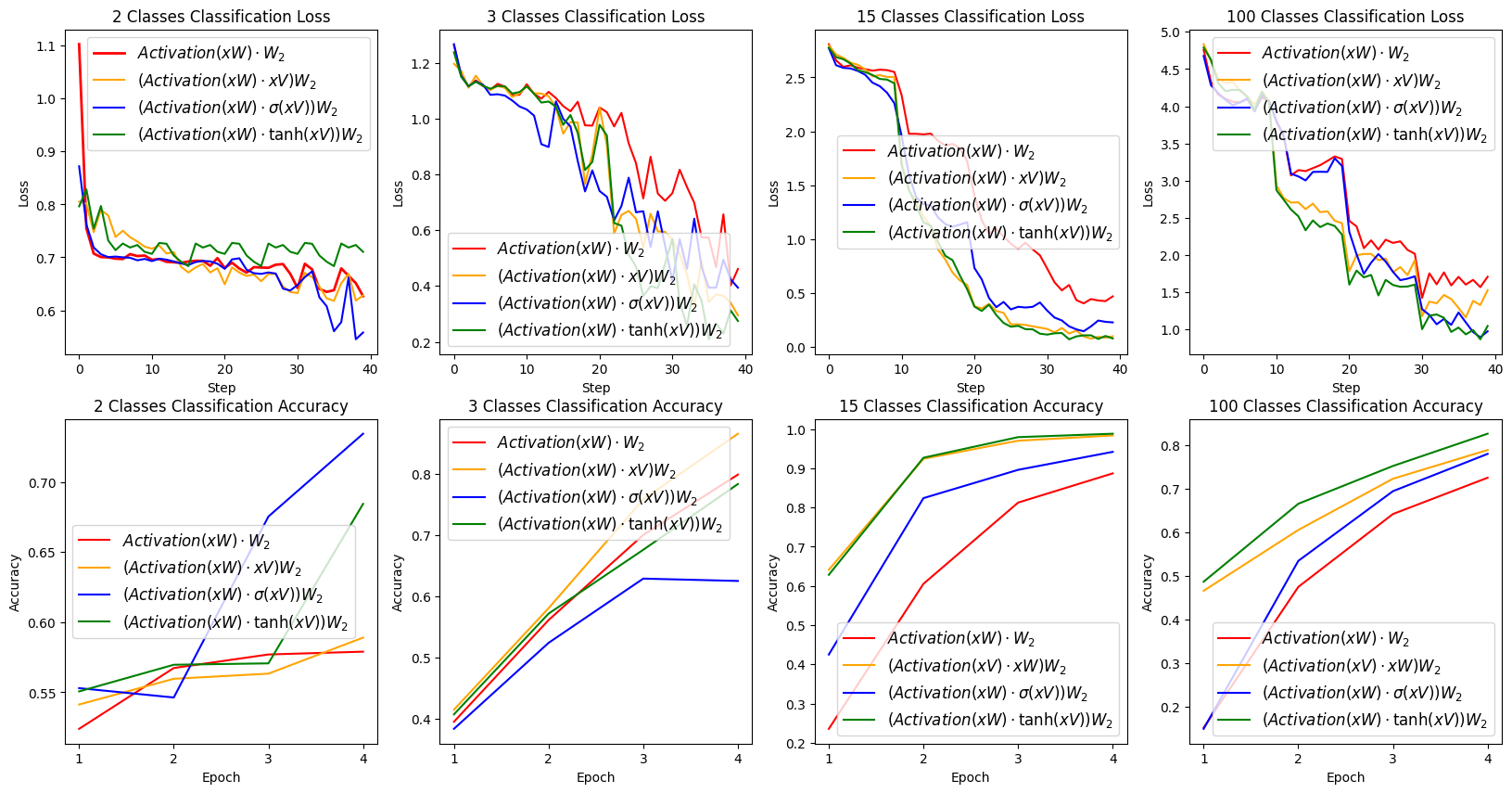}
    \caption{
        (\textbf{Loss and Accuracy of MLP with Different Activation Functions.})
        $FFN_{GLU}$ and $FFN_{DGLU}$ layers both have three weight matrices, while the original $FFN$ layer has only two. To keep the number of parameters and computational complexity unchanged, we reduced the second dimension of $W$ and $V$ of $FFN_{GLU}$ and $FFN_{DGLU}$ layers and the first dimension of $W_2$ to be consistent with the original $FFN$ layer during validation.
    }
    \label{fig:mlp_loss_acc}
\end{figure}

We tried double gated linear units, that is, both linear transformations are passed through activation functions, that is, double gate linear unit (DGLU), and we used two activation functions in the other branch, which are $Sigmod$ and $Tanh$. 

\begin{subequations}
    \begin{align}
        FFN_{SwiSigDGLU}(x, W, V, W_2) &= (Swish(xW) \cdot \sigma(xV))W_2 \\
        \label{eq:ffn_switanhdglu}
        FFN_{SwiTanhDGLU}(x, W, V, W_2) &= (Swish(xW) \cdot \tanh(xV))W_2
    \end{align}
\end{subequations}

The results are shown in Figure~\ref{fig:mlp_loss_acc}.
In the two-category classification task, the loss curve of the $FFN_{DGLU}$ layer using the $\sigma$ function is the lowest, and the accuracy of the $FFN_{DGLU}$ layer using the $\sigma$ function is much higher than other layers. In the three, fifteen, and one hundred category classification tasks, the loss curve of the $FFN_{DGLU}$ layer using the $\tanh$ function is the lowest. In the fifteen category classification task, the accuracy of the $FFN_{DGLU}$ layer using the $\tanh$ function gradually surpasses other activation functions, and the accuracy of the $FFN_{DGLU}$ layer using the $\tanh$ function is the highest in the one hundred category classification task.

In the autoregressive language reasoning task, it can be regarded as a continuous classification task with a category size of the vocabulary size. As the number of categories continues to increase, the loss and accuracy of the $FFN_{DGLU}$ layer using the $\tanh$ function gradually surpass other activation functions. Theoretically, using the $FFN_{SwiTanhDGLU}$ layer will achieve better results in pre-training perplexity and downstream language understanding. Therefore, we use the $FFN_{SwiTanhDGLU}$ layer as the structure of a single expert in the mixture of experts.

\paragraph{Mixture of Experts for Transformers.}
A standard Transformer language model is built by stacking $L$ Transformer blocks, each of which consists of a self-attention block and a feedforward network block, and its output hidden state can be defined as follows:

\begin{subequations}
    \begin{align}
        a^{l}_{1:T} &= \textbf{Self-Attention}(h^{l-1}_{1:T}) + h^{l-1}_{1:T} \\
        ah^{l}_{t} &= \textbf{FFN}(a^{l}_{t}) + a^{l}_{t}
    \end{align}
\end{subequations}

where $T$ is the sequence length, $Self-Attention$ is self-attention, $FFN$ is the feedforward network, $a^{l}_{1:T} \in \R^{T \times d}$ is the hidden state of all tokens after the $l$-th self-attention block, and $ah^{l}_{t} \in \R^{d}$ is the hidden state of the $t$-th token after the $l$-th feedforward network block. For simplicity, we omit normalization in the above formula.

The classic way to construct a mixture of experts language model is to replace the $FFN$ block in the Transformer at specified intervals, where each expert is structurally the same as the standard $FFN$ block, and then assign each token to the topk experts. Replace the $l$-th $FFN$ block with the $MOE$ block, and its output hidden state can be defined as follows:

\begin{subequations}
    \label{eq:attention_moe}
    \begin{align}
        \label{eq:attention_moe:ffn}
        ah^{l}_{t} &= \sum_{i=1}^{N}(g_{i, t} \cdot \textbf{FFN}_{i}(a^{l}_{t})) + a^{l}_{t} \\
        g_{i, t} &= \begin{cases}
            aff_{i, t} & aff_{i, t} \in \text{topK}({aff_{j, t} | 1 \leq j \leq N}, K) \\
            0 & \text{otherwise}
        \end{cases} \\
        aff_{i, t} &= \textbf{Softmax}_{i}(a^{l^{T}}_{t}e^{l}_{i})
    \end{align}
\end{subequations}

where $N$ is the number of experts, $\textbf{FFN}_{i}$ is the feedforward network of the $i$-th expert, $g_{i, t}$ is the gate value assigned to the $i$-th expert for the $t$-th token, $aff_{i, t}$ is the affinity between the token and the expert, $\text{topK}(\cdot, K)$ is the topk selection function of the affinity between the $t$-th token and all experts, and $e^{l}_{i}$ is the centroid of the $i$-th expert in the $l$-th layer. For simplicity, we omit normalization in the above formula.

\paragraph{Mixture of Experts for SSM.}
Following the idea of Mixtures of Experts for Transformers, we can apply it to the state space model.

\begin{subequations}
    \begin{align}
        s^{l}_{1:T} &= \textbf{SSM}(h^{l-1}_{1:T}) + h^{l-1}_{1:T} \\
        sh^{l}_{t} &= \textbf{FFN}(s^{l}_{t}) + s^{l}_{t}
    \end{align}
\end{subequations}

where $T$ is the sequence length, $SSM$ is the state space model, $FFN$ is the feedforward network, $s^{l}_{1:T} \in \R^{T \times d}$ is the hidden state of all tokens after the $l$-th state space model block, and $sh^{l}_{t} \in \R^{d}$ is the hidden state of the $t$-th token after the $l$-th feedforward network block.

Imitating the conventional routing strategy of the mixture of experts language model, we assign each token to the topk experts, replace the $l$-th $FFN$ block with the $MOE$ block, and its output hidden state can be defined as follows:

\begin{subequations}
    \label{eq:ssm_moe}
    \begin{align}
        \label{eq:ssm_moe:ffn}
        sh^{l}_{t} &= \sum_{i=1}^{N}(g_{i, t} \cdot \textbf{FFN}_{i}(s^{l}_{t})) + s^{l}_{t} \\
        g_{i, t} &= \begin{cases}
            aff_{i, t} & aff_{i, t} \in \text{topK}({aff_{j, t} | 1 \leq j \leq N}, K) \\
            0 & \text{otherwise}
        \end{cases} \\
        aff_{i, t} &= \textbf{Softmax}_{i}(s^{l^{T}}_{t}e^{l}_{i})
    \end{align}
\end{subequations}

where $N$ is the number of experts, $\textbf{FFN}_{i}$ is the feedforward network of the $i$-th expert, $g_{i, t}$ is the gate value assigned to the $i$-th expert for the $t$-th token, $aff_{i, t}$ is the affinity between the token and the expert, $\text{topK}(\cdot, K)$ is the topk selection function of the affinity between the $t$-th token and all experts, and $e^{l}_{i}$ is the centroid of the $i$-th expert in the $l$-th layer.

\begin{figure}[H]
    \centering
    \includegraphics[width=1.0\textwidth]{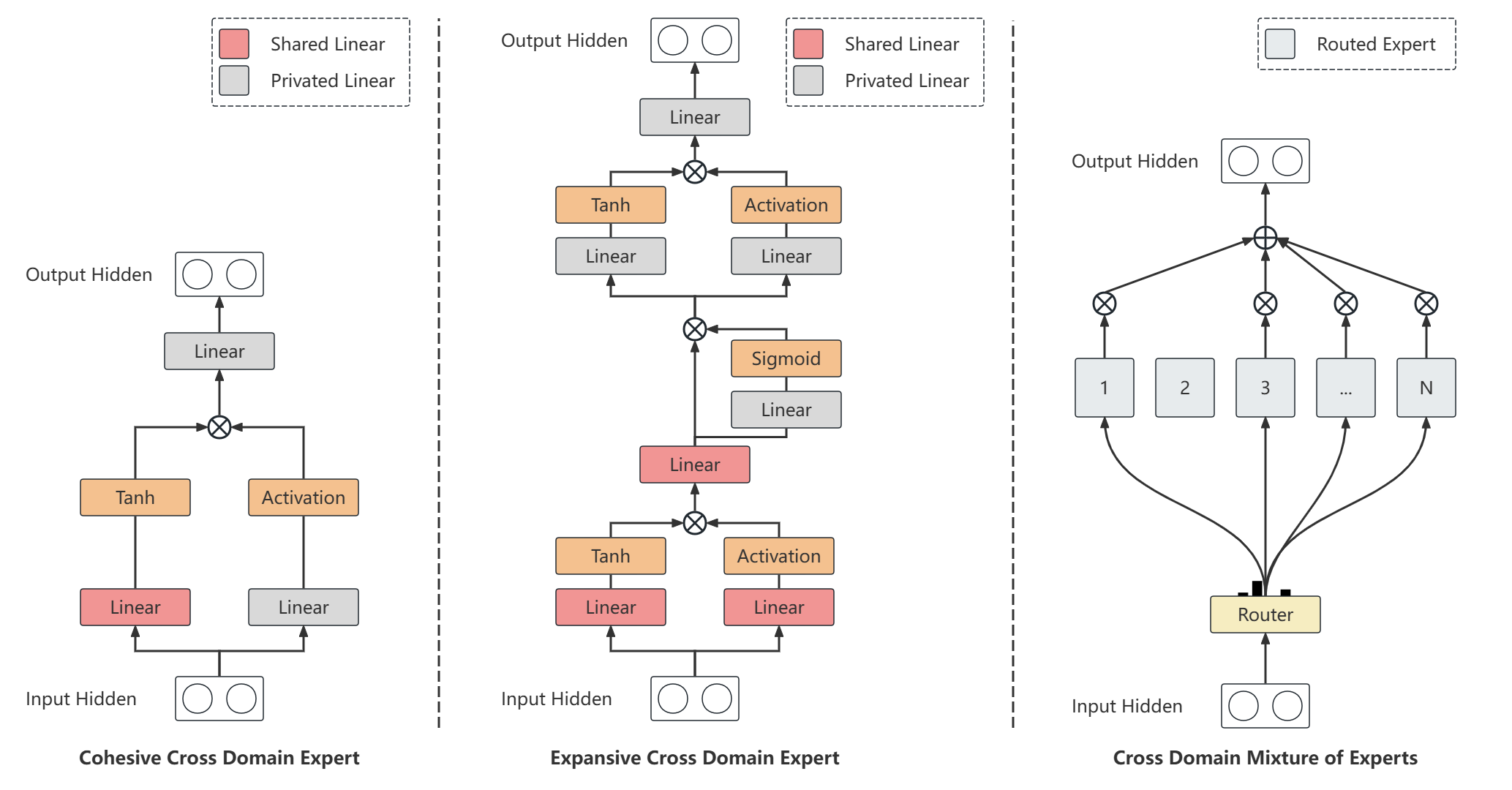}
    \caption{(\textbf{Illustration of Cross Domain Mixture of Experts.}) These three figures respectively show a single cohesive cross domain expert, a single expansive cross domain expert, and the routing expert calculation method, which together form a complete cross domain mixture of experts. It is worth noting that in the specific implementation, Shared Linear can be defined externally and then passed into each private expert, or the shared hidden states can be calculated externally and then passed into each private expert.}
    \label{fig:cdmoe}
\end{figure}

\paragraph{Shared Strategy.}
Under the conventional routing strategy of the mixture of experts, tokens assigned to different experts need common knowledge or information, so multiple routing experts will have redundant parameters that store common information when obtaining their respective parameters. This leads to expert parameter redundancy. However, if these experts have partially shared parameters or the parameters that store common knowledge have already been passed through the parameters, then parameter redundancy can be reduced. The reduction of redundancy will help build a more efficient and specialized mixture of experts.

\paragraph{Cohesive Cross Domain Experts.}
In Attention-MOE~\ref{eq:attention_moe:ffn} and SSM-MOE~\ref{eq:ssm_moe:ffn}, expand $FFN$ into $FFN_{SwiTanhDGLU}$~\ref{eq:ffn_switanhdglu}, which has three weight matrices $W$, $V$, $W_2$. We take matrix $V$ as the shared parameter of different $FFN$ in the mixture of experts, and take $W$ and $W_2$ as the private parameters of each $FFN$. Re-represent the parameters as $W^{p}$, $V^{s}$, $W_2^{p}$, where $p$ is the private parameter and $s$ is the shared parameter, and keep other parameters and formulas unchanged.

\begin{align}
    FFN^{Cohesive}(x, W^{p}, V^{s}, W_2^{p}) &= (Swish(xW^{p}) \cdot \tanh(xV^{s}))W_2^{p}
\end{align}

This way of sharing parameters inside $FFN$ can reduce parameter redundancy and the total number of parameters. We call it cohesive cross domain experts. However, this method is only suitable for cases with a small number of experts, because the shared parameter $V^{s}$ cannot adjust the number of parameters independently and must be consistent with $W^{p}$, which will cause the proportion of shared parameters to decrease as the number of experts increases.

\paragraph{Expansive Cross Domain Experts.}
To cope with a large number of experts while maintaining the proportion of shared parameters unchanged, still taking $FFN_{SwiTanhDGLU}$~\ref{eq:ffn_switanhdglu} as an example, use $p$ to represent private parameters, $s$ to represent shared parameters, define $W^{p}$, $V^{p}$, $W_2^{p}$ as private $FFN$ parameters, $W^{s}$, $V^{s}$, $W_2^{s}$ as shared $FFN$ parameters, and $W_3^{p}$ as the gate parameter that receives shared hidden states to private parameters in each private $FFN$. Re-represent the Attention-MOE~\ref{eq:attention_moe:ffn} and SSM-MOE~\ref{eq:ssm_moe:ffn} as expansive cross domain experts, keeping other parameters and formulas unchanged.

\begin{subequations}
    \begin{align}
        FFN^{p}(x, W^{p}, V^{p}, W_2^{p}) &= (Swish(xW^{p}) \cdot \tanh(xV^{p}))W_2^{p} \\
        FFN^{s}(x, W^{s}, V^{s}, W_2^{s}) &= (Swish(xW^{s}) \cdot \tanh(xV^{s}))W_2^{s} \\
        FFN^{Expansive}(x, W_3^{p}) &= FFN^{p}(FFN^{s}(x) \cdot FFN^{s}(x)W_3^{p}) \\
        ah^{l}_{t} &= \sum_{i=1}^{N}(g_{i, t} \cdot FFN^{Expansive}_{i}(s^{l}_{t})) + s^{l}_{t} \\
        sh^{l}_{t} &= \sum_{i=1}^{N}(g_{i, t} \cdot FFN^{Expansive}_{i}(s^{l}_{t})) + s^{l}_{t}
    \end{align}
\end{subequations}

In the expansive cross domain experts, tokens routed to various experts first pass through the shared $FFN$ parameters that store common knowledge, and then pass through the gate of the private $FFN$ to let the private experts decide which information from the shared hidden states they should receive. Compared with the shared expert isolation~\ref{fig:shared_expert_isolation}, the expansive cross domain experts achieve more accurate and targeted knowledge acquisition capabilities, and retain the advantage of being able to adjust the proportion of shared parameters arbitrarily.

\subsection{Architecture Design}
\label{sec:architecture_design}

\paragraph{Self-Attention's Characteristics and Drawbacks.}
Let's review the algorithm of standard self-attention~\ref{alg:self_attention}, Self-attention's parameters include the weight matrices $W_Q$, $W_K$, $W_V$ and bias of $Q$, $K$, $V$, and the complete self-attention layer also includes an output weight matrix $W_O$ and bias.

First, calculate $Q$, $K$, $V$ that is $Q = xW_Q$, $K = xW_K$, $V = xW_V$, the input and output shapes of this matrix multiplication are $[B, L, D] \times [D, D] = [B, L, D]$, the computational complexity is $3 \times 2BLD^2 = 6BLD^2$.

Then calculate the matrix multiplication of $QK^T$, the input and output shapes are $[B, L, D] \times [B, D, L] = [B, L, L]$, the computational complexity is $2BL^2D$.

Then calculate the weighted sum on $V$, the input and output shapes of this matrix multiplication are $[B, L, L] \times [B, L, D] = [B, L, D]$, the computational complexity is $2BL^2D$.

Finally, it is the linear mapping after self-attention, the input and output shapes of this matrix multiplication are $[B, L, D] \times [D, D] = [B, L, D]$, the computational complexity is $2BLD^2$.

In summary, the computational complexity of the self-attention layer is $8BLD^2 + 4BL^2D$, the total computational complexity of the multi-head self-attention variant is the same, but it can be calculated in parallel when calculating $(QK^T)V$, which is faster.

Next, we can calculate the intermediate activation memory usage of the self-attention layer $x_{out} = softmax(\frac{QK^T}{\sqrt{d}})V \cdot W_O + x$.

For $Q$, $K$, $V$, we need to save their common input $x$, the size of $x$ is $[B, L, D]$, the number of elements is $BLD$, the memory usage is $2BLD$.

For the matrix multiplication of $QK^T$, we need to save $Q$, $K$, the shapes of these two tensors are $[B, L, D]$, the memory usage is $4BLD$.

For $softmax(\frac{QK^T}{\sqrt{d}})$, we need to save the input of the function $QK^T$, the memory usage is $2BL^2h$, where $h$ is the number of heads of multi-head attention, the shape of $Q$ is $[B, head\_num, L, per\_head\_dim]$, the shape of $K$ is $[B, head\_num, per\_head\_dim, L]$, the shape of $QK^T$ is $[B, head\_num, L, L]$, after calculating $softmax$, a $dropout$ operation will be performed, we need to save a $mask$ matrix, the shape of the $mask$ matrix is the same as $QK^T$, the memory usage is $BL^2h$.

Calculate the attention on $V$, that is $AV$, we need to save $A$, the memory usage is $2BL^2h$, and $V$, the memory usage is $2BLD$, the memory usage is $2BL^2h + 2BLD$.

Calculate the final output mapping and a $dropout$ operation, the output mapping needs to save its input $y$, the memory usage is $2BLD$, the $dropout$ operation needs to save a $mask$ matrix, the memory usage is $BLD$, the memory usage is $2BLD + BLD = 3BLD$.

Therefore, by adding the memory usage of all the intermediate activations above, the memory usage of the intermediate activations of the self-attention layer is $11BLD + 5BL^2h$.

Through the above calculation, we can see that the computational complexity and storage complexity of self-attention increase quadratically with the sequence length $L$, which limits the maximum sequence length $L$,
but we can easily see two characteristics of self-attention from the formula:
\begin{itemize}
    \item Self-attention does not compress context information, that is, it can capture the dependency relationship between any position elements in the sequence, regardless of distance.
    \item Self-attention weights each element, each element considers the information of all other elements in the sequence, and produces a context-aware state.
\end{itemize}

\paragraph{SSM's Characteristics and Drawbacks.}
The computational complexity of SSM increases linearly with the sequence length $L$, and we will not go into details here.
Let's review the Selective State Model~\ref{sec:background:ssm}, which has two important characteristics:
\begin{itemize}
    \item The data-dependent $\dB$ matrix more finely controls whether to let the input $x$ affect the state $h$, the data-dependent $\dA$ matrix controls the influence from the previous state $h_{t-1}$ to the current state $h_t$, and the data-dependent $\C$ matrix controls whether the state $h$ can affect the output $y$, the data dependency of these three matrices gives SSM the ability to selectively process information.
    \item Because the number of parameters in the word embedding matrix is also relatively large, the linear recursive aggregation state assigned to the last token allows us to calculate the matrix multiplication of $[B, D] \times [D, V] = [B, V]$ to obtain $logits$.
\end{itemize}

Of course, the drawbacks of SSM are also obvious, 
by approximating the optimal solution of the state matrix $A$ through functions to remember history,
it can only accurately capture the state change from $h_{t-1}$ to $h_t$,
gradually decay the old state information,
which leads to SSM performing poorly in tasks that require strong copying or context learning ability or long context reasoning.

\paragraph{Combination of Attention and SSM.}
If a human who loves Shiba Inu reads an article, we can divide it into four stages: observation, thinking, conception, and expression.

\begin{figure}[H]
    \centering
    \includegraphics[width=0.4\textwidth]{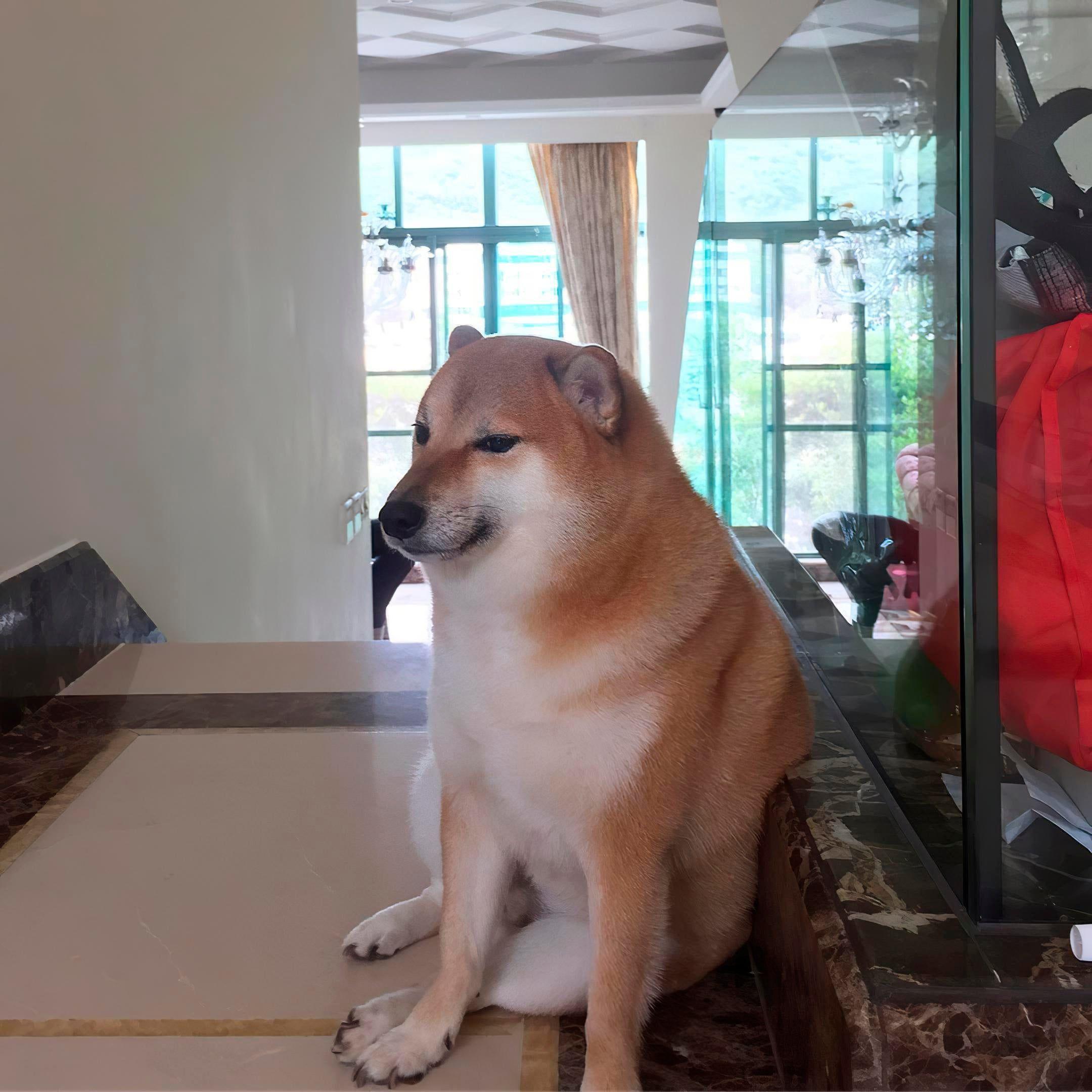}
    \caption{
        \textbf{Cheems.} 
        The prototype of the Shiba Inu Cheems that became popular on the Internet. If you are a human who has been stamped with the idea of loving Shiba Inu (uncompressed memory dependency of any length), you should have a sudden realization --- the process of observation, thinking, conception, and expression corresponds to the combination of SSM and self-attention in the Observer, Thinker, Conceiver, and Expresser modules of the OTCE architecture.
    }
    \label{fig:cheems}
\end{figure}

In the observation stage, we are often attracted by the pictures first, and this interesting picture priority corresponds to the selective processing ability of SSM for information.

In the thinking stage, we carefully observe the decadent expression and sitting posture of the creature in the picture to think about the relevance to the love of Shiba Inu, this taking the current information and past memory to build long-term dependencies corresponds to the global dependency of self-attention.

In the conception stage, we combine the current information with relevant past memories to deduce the logical relationship step by step and finally draw a conclusion, which corresponds to the linear recursive aggregation state of SSM.

In the expression stage, we accurately say the name found from the distant memory: "This is (surprised) Cheems (love)!", which corresponds to the context-aware state of self-attention considering all elements.

This example is not very appropriate, it is more appropriate to compare it with looking up in the phone book, but under our strong desire, we finally chose to compare it with the Internet celebrity dog Cheems~\ref{fig:cheems}.
These four stages correspond to the combination of SSM and self-attention in the Observer, Thinker, Conceiver, and Expresser modules of the OTCE architecture~\ref{fig:otce}.

\begin{figure}[H]
    \centering
    \includegraphics[width=1.0\textwidth]{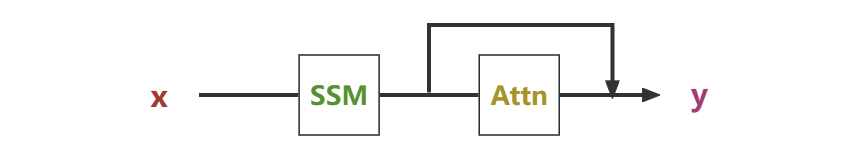}
    \caption{
        \textbf{Expresser.}
        The Expresser module reweights the information with gradually decaying long-term dependency relationships through re-attention to all elements, and learns how SSM recursively aggregates information to correct knowledge routing bias.
        Maybe we can do something on the branch that skips the connection attention.
    }
    \label{fig:expresser}
\end{figure}

It is worth noting that we design the Expresser module~\ref{fig:expresser} to consider the information of all elements of the linear recursive aggregation state of the previous layer SSM after weighting with self-attention and then add it to it, and finally obtain the final output through a combination of a linear layer, an activation function, and another linear layer.
The purpose of doing this is to prevent the bias caused by too many SSM layers after the Thinker module, and at the same time, it can retain the information of the aggregation state for efficient reasoning.

\paragraph{Combination of MLP and MOE.}
After ablation experiments, we found that using MLP in the first layer and the last layer of the model performs best on the pre-training perplexity.
Since the self-attention part of the Expresser considers the information of all elements of the linear recursive aggregation state of the previous layer SSM after weighting and then adds it to it, we also consider it as the last layer, that is, the first layer of the Observer module and the last layer of the Conceiver and Expresser modules use MLP, and other layers use the combination of MOE is the best.
Ablation experiments will be detailed in the following sections.
  \section{Empirical Validation}
\label{sec:experiments}

We empirically validated OTCE on semantic similarity, long-short text classification, natural language inference, keyword recognition, different domain selection questions, context learning, and multi-query associative recall tasks.
The importance of re-attention weighting in the Expresser module before output is verified in Section~\ref{sec:attention_before_output_is_important},
and the use of positional encoding in SSM and Attention to connect the token positions in the hybrid architecture is verified in Section~\ref{sec:positional_encoding_that_connected_ssm_to_attention}.
The ablation of different combinations of MOE and MLP in the OTCE architecture, and the comparison of MOE with different shared parameter strategies in pre-training, fine-tuning, and multi-domain selection question tasks are verified in Section~\ref{sec:cross_domain_improve_efficiency},
and the verification on context learning and multi-query associative recall tasks is verified in Section~\ref{sec:associative_recall}.

\subsection{Training Dataset and Hyperparameters}
\label{sec:training_dataset_and_hyperparameters}

The pre-training dataset is a mixture of open-source datasets including Book, Wikipedia, UNCorpus, translation2019zh, WikiMatrix, news-commentry, ParaCrawl v9, ClueCorpusSmall, CSL, news-crawl, etc., and the SFT fine-tuning dataset is 2 million instructions from the Belle dataset.

The learning rate formula comes from Attention is All You Need~\citep{transformer2017}, to pay tribute to the first paper we read as a sophomore ordinary student starting from scratch to study machine learning.

\begin{equation*}
    lrate = d_{model}^{-0.5} \times \min(step\_num^{-0.5}, step\_num \times warmup\_steps^{-1.5})
\end{equation*}

In addition, we did not use linear bias terms, the warm-up steps are $10\%$ of the total steps, $RMSNorm$ is used instead of $LayerNorm$, $AdamW$ hyperparameters are set to $\beta_1 = 0.9, \beta_2 = 0.98, \epsilon = 10^{-6}$, and GLM~\citep{du2022glm} is used as the architecture backbone, that is, bidirectional GLM universal language model: MASK spans in the training text sequence, with autoregressive filling as the training target, requiring the model to predict the masked token.
The validation environment is the open-source PyTorch image provided by Nvidia~\citep{pytorch}.

\subsection{Attention before Output is Important}
\label{sec:attention_before_output_is_important}

In this section, we will not adopt most of the improvements mentioned earlier, but focus on verifying the importance of attention weighting the recursively aggregated state of SSM at the output stage.
As shown in Table~\ref{tab:OTCE_vs_Jamba_Params}, when the number of layers is 8, one of the combinations of OTCE is very similar to a single Jamba block of another hybrid architecture Jamba~\citep{lieber2024jamba}.
The main difference between the two is only in the distribution strategy of MLP and MOE, and the last layer of OTCE is the Expresser module.

\begin{table}[H]
    \footnotesize
    \centering
    \caption{
        (\textbf{Jamba and OTCE Parameters}.)
        The parameters of Jamba and OTCE with 8 layers, the positional information comes from the convolution before each SSM, without RoPE, and without MOE, the difference is only in the last layer module, Jamba is SSM + MLP, OTCE is Attention + MLP.
    }
    \label{tab:OTCE_vs_Jamba_Params}
    \begin{tabular}{@{}lllllllll@{}}
    \toprule
    \sc{Model} & \sc{Params} & \sc{Layers} & \sc{Model Dim} & \sc{FFN Dim} & \sc{Attn Heads} & \sc{Group} & \sc{State Dim} & \sc{Hybrid (S=SSM, A=Attention, M=MLP)} \\
    \midrule
    \label{tab:280M_Jamba_Params}
    Jamba & 283M & 8 & 1024 & 4096 & 16 & 8 & 16 & SMSMSMSMAMSMSMSM \\
    \label{tab:280M_OTCE_Params}
    OTCE & 281M & 8 & 1024 & 4096 & 16 & 8 & 16 & SMSMSMSMAMSMSMAM \\
    \bottomrule
    \end{tabular}
\end{table}

\begin{table}[H]
    \centering
    \caption{
        (\textbf{Jamba and OTCE Results}.)
        The perplexity performance of Jamba and OTCE in the pre-training stage is very close, but in the SFT fine-tuning stage, the perplexity of OTCE is significantly lower than Jamba.
        We select the evaluations for the CLUEbenchmark~\citep{xu-etal-2020-clue} that are more suitable for the current parameter volume:
        AFQMC: semantic similarity, judge whether two sentences have the same meaning.
        TNEWS: short text classification, from the news section of today's headlines, judge the category of the news.
        IFLYTEK: long text classification, a long text dataset about software descriptions, judge the application theme of the software description.
        CMNLI: language reasoning, judge the logical relationship between two sentences.
        WSC: pronoun disambiguation, judge which noun the pronoun in the sentence refers to.
        CSL: keyword recognition, taken from the abstract and keywords of the paper, judge whether the keyword is in the abstract.
        OTCE maintains a slight lead in AFQMC, TNEWS, IFLYTEK, WSC, and CSL, and slightly lags behind in CMNLI, with an average accuracy improvement of $4.25\%$ over Jamba.
    }
    \label{tab:OTCE_vs_Jamba_Results}
    \begin{tabular}{@{}llllllllll@{}}
    \toprule
    \sc{Model} & \sc{PreTrain} & \sc{FineTune} & \sc{AFQMC} & \sc{TNEWS} & \sc{IFLYTEK} & \sc{CMNLI} & \sc{WSC} & \sc{CSL} & \sc{Average} \\
    & \sc{ppl $\downarrow$} & \sc{ppl $\downarrow$} & \sc{acc $\uparrow$} & \sc{acc $\uparrow$} & \sc{acc $\uparrow$} & \sc{acc $\uparrow$} & \sc{acc $\uparrow$} & \sc{acc $\uparrow$} & \sc{acc $\uparrow$} \\
    \midrule
    Jamba & 2.53 & 1.98 & 78.65 & 63.15 & 72.40 & 79.66 & 76.64 & 84.83 & 75.89 \\
    OTCE & 2.51 & 1.69 & 79.48 & 77.98 & 72.71 & 79.59 & 79.93 & 85.06 & 79.12 \\
    \bottomrule
    \end{tabular}
\end{table}

As shown in Table~\ref{tab:OTCE_vs_Jamba_Results}, the difference in perplexity between Jamba and OTCE in the pre-training stage is negligible, which indicates that both have learned almost the same knowledge from pre-training. However, in the SFT fine-tuning stage, the model usually needs to remember specific formats based on the context and reason based on these formats in the subsequent context. This has been proven to be a deficiency in the ability of SSM to aggregate state information recursively. The Attention part in a single Jamba block is embedded between the SSM layers, which leads to the ability of attention to directly route the knowledge of each answer to a single answer token being attenuated by the subsequent three SSM layers, resulting in a bias in knowledge routing.

The Expresser module of OTCE re-weights the state information recursively aggregated by SSM with attention, which is equivalent to learning and weighting how the information from the first token to the last token is recursively aggregated. Then, the two types of state information are mixed together, effectively avoiding the bias in knowledge routing. Therefore, we can now explain why OTCE has a significant advantage in most validation tasks.

For tasks such as AFQMC, which judge whether the meanings of sentences are the same, CSL, which judge whether keywords are in the abstract rather than extracting keywords from the abstract, and CMNLI, which judge whether the relationship between two sentences is entailment, neutral, or contradiction, they mainly rely on selectively extracting key information on the sequence and recursively aggregating state information. Therefore, the performance gap between Jamba and OTCE on these tasks is not significant, and the reason why OTCE slightly outperforms Jamba in AFQMC and CSL may be that it corrects some knowledge routing biases.

In tasks such as TNEWS, which predict the category of short sentences, since the sentences do not contain or contain very little context information, they are very dependent on the quality of pre-training and fine-tuning. This requires that the knowledge learned in the pre-training stage can be correctly and accurately guided in the fine-tuning stage, which leads to Jamba's performance on TNEWS being far worse than OTCE.

In contrast to TNEWS, the text in the IFLYTEK task is all long context, and the vocabulary of the application theme appears multiple times. This constitutes a relatively simple single-query associative recall task, where the vocabulary of the single main theme that appears multiple times is allowed to be stored in the state by the selective state space model and becomes the main factor affecting the output. Therefore, the performance gap between the two on IFLYTEK is not significant.

In the WSC task, which is a pronoun disambiguation task, the model needs to be able to trace back to the noun that appeared before the pronoun after the pronoun appears and point the pronoun to the correct noun. This belongs to the additional requirement of understanding the current language in the single-query associative recall task. Therefore, the reason why the performance gap between the two on WSC is second in these tasks is obvious, with the two main factors being the difference in fine-tuning quality and the re-attention of the recursive aggregation information in the Expresser module of OTCE on the long-term strong dependency relationship.

\subsection{Positional Encoding that connected SSM to Attention}
\label{sec:positional_encoding_that_connected_ssm_to_attention}

In this section, we evaluate the positional encoding strategy that connects SSM to Attention.
The parameter settings used are consistent with the 281M OTCE model in Table~\ref{tab:280M_OTCE_Params}.

We conducted four pre-training experiments, namely without using RoPE, only using Attention RoPE, only using SSM RoPE, and using both Attention and SSM RoPE.
The experiments used a sequence length of 8192 and a RoPE base frequency of 10000.
The experimental results are detailed in Table~\ref{tab:positional_encoding_that_connected_ssm_to_attention}.

\begin{table}[H]
    \centering
    \caption{
        (\textbf{Evaluation of Different Position Information Sources}.)
        Pre-training perplexity performance without RoPE, only with Attention RoPE, only with SSM RoPE, and with both Attention and SSM RoPE, with Attention accounting for $12.5\%$, SSM accounting for $37.5\%$, and MLP accounting for $50\%$.
    }
    \label{tab:positional_encoding_that_connected_ssm_to_attention}
    \begin{tabular}{@{}llllllllll@{}}
    \toprule
    \sc{Non RoPE} & \sc{Only Attention RoPE} & \sc{Only SSM RoPE} & \sc{Both Attention and SSM RoPE} \\
    \sc{ppl $\downarrow$} & \sc{ppl $\downarrow$} & \sc{ppl $\downarrow$} & \sc{ppl $\downarrow$} \\
    \midrule
    2.51 & 2.49 & 2.26 & 2.10 \\
    \bottomrule
    \end{tabular}
\end{table}

From the experimental results, we can observe that there is little difference in perplexity performance between not using RoPE and only using Attention RoPE.
This finding is consistent with recent large-scale studies on hybrid architectures~\citep{waleffe2024mamba-study}.

However, when only using SSM RoPE, the perplexity decreases by nearly $10\%$ compared to not using RoPE.
Moreover, when using both Attention and SSM RoPE, the perplexity decreases by nearly $7\%$ compared to only using SSM RoPE.

This indicates that the one-dimensional convolution layer before SSM provides local positional information in the model's hidden dimension, promoting the consideration of adjacent positional information during linear recursive state updates, which is also applicable to the matrix $D$ that skips the connection between input and output in SSM.
After selective filtering by SSM, matrix $D$ reassigns discrete effective information with continuous positional context.
However, the local positional information provided by the one-dimensional convolution has limited impact on building dependencies between any two elements in a sequence of any length in Attention.
Therefore, when only using Attention RoPE, although the improvement in perplexity is not significant compared to not using RoPE, there is still an improvement.
We believe that the slight negative reason for the improvement is that Attention accounts for a small proportion of the entire model, while the positive reason is that the final Expresser module combines Attention with MLP.
The negative contrast of this observation is also reflected in the recent large-scale study on hybrid architectures~\citep{waleffe2024mamba-study}, which caused a negative impact on a sequence length of 16K.

When only using SSM RoPE, we are actually injecting continuous global relative positional information into the SSM state.
Even after selective filtering, this information is still stored in the SSM state in discrete positional form, affecting the linear recursive $h_t$ to consider the discrete relative position of the previous effective information state.
In the subsequent Attention calculation, it still relies on the local positional information provided by the one-dimensional convolution in the hidden dimension.
Nevertheless, when only using SSM RoPE, the perplexity performance decreases by nearly $10\%$ compared to not using RoPE, significantly better than not using RoPE.
This result indicates that performing position calculation and update in the SSM state is far superior to the method of providing local positional information by convolution first and then reassigning continuous positional information by matrix $D$.

When using both Attention and SSM RoPE, we can utilize the global relative positional information in the SSM state.
After selective filtering by SSM, discrete effective information regains continuous positional information in Attention RoPE.
Through Attention calculation, the model can not only consider the relative position between effective information states and ineffective information states but also build dependencies between any two elements in a sequence of any length and any element.
This is why when using both Attention and SSM RoPE, the perplexity continues to decrease by nearly $7\%$ compared to only using SSM RoPE.
Moreover, it is worth noting that in this case, we can completely remove the one-dimensional convolution layer in the Mamba structure and the matrix $D$ in SSM, saving much more computation than the additional computation required for RoPE.

\subsection{Cross-Domain can Improve the Efficiency of Fine-tuning}
\label{sec:cross_domain_improve_efficiency}

In this section, we will conduct a combination ablation experiment of MOE and MLP in the OTCE architecture, and evaluate the zero-shot and five-shot accuracy of shared expert isolation mixed expert models, cohesive cross domain mixed expert models, and expansive cross domain mixed expert models in pre-training, fine-tuning, and cross-domain multiple-choice questions.

The parameters of the MOE and MLP combination ablation experiment are set as shown in Table~\ref{tab:MOE_vs_MLP_Params}, and the results are shown in Table~\ref{tab:MOE_vs_MLP_Results}.

\begin{table}[H]
    \centering
    \caption{
        (\textbf{The Parameters of MOE and MLP Combination Ablation Experiment}.)
        Because the ablation experiment is very time-consuming, we set a small parameter size to facilitate the ablation experiment with limited computing resources.
    }
    \label{tab:MOE_vs_MLP_Params}
    \begin{tabular}{@{}llllllll@{}}
    \toprule
    \sc{Layers} & \sc{Model Dim} & \sc{Attn Heads} & \sc{Group} & \sc{State Dim} & \sc{Experts} & \sc{Experts topK} \\
    \midrule
    8 & 1024 & 16 & 2 & 16 & 4 & 2 \\
    \bottomrule
    \end{tabular}
\end{table}

\begin{table}[H]
    \centering
    \caption{
        (\textbf{The Results of MOE and MLP Combination Ablation Experiment}.)
        We temporarily use ordinary MOE~\ref{eq:attention_moe}~\ref{eq:ssm_moe}, 
        adjust the dimension size of $FFN$ to ensure that any combination has the same total parameters, 
        test several representative MOE and MLP combinations, 
        iterate the same number of steps on the pre-training subset, 
        and evaluate the performance of different combinations through perplexity.
    }
    \label{tab:MOE_vs_MLP_Results}
    \begin{tabular}{@{}lllllll@{}}
    \toprule
    \sc{Hybrid (S=SSM, A=Attention, M=MLP, E=MOE)} & \sc{MOE Ratio} & \sc{ppl $\downarrow$} \\
    \midrule
    \SM \SM \SM \SM \AM \SM \SM \AM & 0\% & 4.86 \\
    \SE \SM \SM \SM \AM \SM \SM \AM & 6.25\% & 4.83 \\
    \SM \SM \SM \SM \AE \SM \SM \AM & 6.25\% & 4.84 \\
    \SM \SM \SM \SM \AM \SE \SM \AM & 6.25\% & 4.82 \\
    \SM \SM \SM \SM \AM \SM \SE \AM & 6.25\% & 4.88 \\
    \SM \SM \SM \SM \AM \SM \SM \AE & 6.25\% & 4.92 \\
    \SE \SM \SM \SM \AM \SM \SM \AE & 12.5\% & 4.68 \\
    \SM \SM \SM \SE \AM \SE \SM \AM & 12.5\% & 4.68 \\
    \SM \SM \SE \SE \AM \SM \SM \AM & 12.5\% & 4.67 \\
    \SM \SM \SM \SM \AM \SE \SE \AM & 12.5\% & 4.68 \\ 
    \SM \SE \SE \SE \AM \SM \SM \AM & 18.75\% & 4.77 \\
    \SE \SM \SM \SM \AM \SM \SE \AE & 18.75\% & 4.69 \\
    \SM \SM \SM \SE \AE \SE \SM \AM & 18.75\% & 4.63 \\
    \SM \SE \SM \SE \AM \SE \SM \SE & 25\% & 4.61 \\
    \SM \SE \SE \SE \AE \SE \SM \AM & 31.25\% & \textbf{4.48} \\
    \SM \SE \SE \SE \AE \SE \SE \AM & 37.5\% & 4.56 \\
    \SE \SE \SE \SE \AE \SE \SE \AE & 50\% & 4.54 \\
    \bottomrule
    \end{tabular}
\end{table}

In most cases, the perplexity of different combinations decreases with the increase of the MOE ratio, and the training cost of MOE is lower than that of MLP with the same total parameters, so we choose the combination with a higher MOE ratio as much as possible.
We found that the perplexity is the lowest when the first layer and the last layer (because of the special nature of the Expresser module, it should be regarded as the last two layers) of the model use MLP, and we speculate that using dense activation parameters in the two parts of the model that process the input and prepare the output can help improve stability.
So we choose the first layer of the Observer module as MLP, the last layer of the Conceiver module as MLP, and the last combination of the Expresser module as MLP, as shown in Figure~\ref{fig:otce}.

\begin{table}[H]
    \footnotesize
    \centering
    \caption{
        (\textbf{The Parameters of Three Mixed Expert Models with Shared Parameters}.)
        We mainly increase the number of model layers to improve the data saturation limit of the model, 
        and abbreviate Shared Expert Isolation Mixture of Experts as SEIMOE, 
        Cohesive Cross Domain Mixture of Experts as CCDMOE, 
        Expansive Cross Domain Mixture of Experts as ECDMOE.
        In addition, because the shared parameter ratio of the cohesive cross domain expert is only related to the number of experts and cannot be adjusted independently, 
        we first determine the total number of parameters and the shared parameter ratio between experts of the cohesive cross domain mixed expert model, 
        and then adjust the size of the $FFN$ of the shared expert isolation mixed expert model and the expansive cross domain mixed expert model to ensure that the total number of parameters and the shared parameter ratio between experts of the three models are the same.
    }
    \label{tab:three_moe_params}
    \begin{tabularx}{0.9\linewidth}{@{}cccccc@{}}
    \toprule
    \sc{Layers} & \sc{Model Dim} & \sc{Attn Heads} & \sc{Group} & \sc{State Dim} & \sc{Hybrid} \\
    & & & & & \sc{s=ssm a=attention m=mlp e=moe} \\
    \midrule
    24 & 1024 & 16 & 2 & 64 & $\SM + \SE \times 11 + \AE \times 2 + \SE \times 8 + \SM + \AM$ \\
    \bottomrule
    \end{tabularx}

    \vspace{0em}

    \begin{tabularx}{0.9\linewidth}{@{}ccccccc@{}}
    \toprule
    \sc{Model} & \sc{Parameters} & \sc{Experts} & \sc{Experts topK} & \sc{Shared parameter ratio} & \sc{PreTrain} & \sc{FineTune} \\
    & & & & & \sc{ppl $\downarrow$} & \sc{ppl $\downarrow$} \\
    \midrule
    SEIMOE & 1B & 4 & 2 & 1:8 & 1.99 & 1.55 \\
    CCDMOE & 1B & 4 & 2 & 1:8 & 1.93 & 1.52 \\
    ECDMOE & 1B & 4 & 2 & 1:8 & 1.88 & 1.38 \\
    \bottomrule
    \end{tabularx}
\end{table}

We list the basic parameter settings of the shared expert isolation mixed expert model, cohesive cross domain mixed expert model, and expansive cross domain mixed expert model in Table~\ref{tab:three_moe_params}, the hybrid of SSM, Attention, MLP, and MOE, the total number of parameters, the number of experts, the number of expert activation, and the shared parameter ratio between experts are the same.

For sparse activation model architectures, an important criterion for evaluating whether they are excellent is the data efficiency of training, as shown in Table~\ref{tab:three_moe_params}, the comparison of pre-training perplexity and fine-tuning perplexity of SEIMOE, CCDMOE, and ECDMOE, when we know that our training data volume cannot train the model to the data saturation limit, we set the total number of parameters and the shared parameter ratio to be the same, so we can see the training data efficiency of the model from the perplexity.
The pre-training perplexity of CCDMOE is slightly better than SEIMOE, and the pre-training perplexity of ECDMOE is slightly better than CCDMOE, but in the fine-tuning perplexity performance, ECDMOE is far better than SEIMOE and CCDMOE, with an increase of $10.96\%$ and $9.21\%$, respectively.

In Table~\ref{tab:three_moe_evaluations}, we detail the zero-shot and five-shot scores of SEIMOE, CCDMOE, and ECDMOE in each domain.
The average scores of CCDMOE and ECDMOE on zero-shot and five-shot are better than SEIMOE, and the average scores of ECDMOE on zero-shot and five-shot are better than CCDMOE.
It is worth noting that the average scores of SEIMOE and CCDMOE after Five-shot Fine-tune are lower than the zero-shot scores, while ECDMOE performs well, with an increase of $2\%$, which is consistent with the performance of the three models in fine-tuning perplexity, indicating that ECDMOE is more efficient in training data than SEIMOE and CCDMOE.

\begin{table}[!ht]
    \footnotesize
    \centering
    \caption{
        (\textbf{The Zero-shot and Five-shot Evaluations of Three Mixed Expert Models with Shared Parameters}.)
        These tasks come from CEvalBenchmark~\citep{huang2023ceval}, 
        we detail the zero-shot and five-shot accuracy of three mixed expert models with shared parameters in each subtask.
    }
    \label{tab:three_moe_evaluations}
    \begin{tabular}{@{}lllllll@{}}
    \toprule
    \sc{Task} & \sc{SEIMOE} & \sc{CCDMOE} & \sc{ECDMOE} & \sc{SEIMOE} & \sc{CCDMOE} & \sc{ECDMOE} \\
    & \sc{zero-shot $\uparrow$} & \sc{zero-shot $\uparrow$} & \sc{zero-shot $\uparrow$} & \sc{five-shot $\uparrow$} & \sc{five-shot $\uparrow$} & \sc{five-shot $\uparrow$} \\
    \midrule
    computer network & 42.10 & 52.63 & 53.51 & 47.37 & 52.63 & 54.87 \\
    operating system & 63.16 & 52.63 & 60.11 & 47.37 & 47.36 & 51.41 \\
    computer architecture & 61.90 & 52.38 & 54.21 & 61.90 & 52.38 & 52.09 \\
    college programming & 27.03 & 40.54 & 36.36 & 35.14 & 32.43 & 36.36 \\
    college physics & 36.84 & 36.84 & 37.65 & 42.11 & 42.11& 43.15 \\
    college chemistry & 33.33 & 45.83 & 44.37 & 37.50 & 54.17 & 61.34 \\
    advanced mathematics & 36.84 & 47.37 & 43.76 & 42.11 & 57.89 & 47.76 \\
    probability and statistics & 44.44 & 44.44 & 45.66 & 44.44 & 38.89 & 41.21 \\
    discrete mathematics & 56.25 & 43.75 & 44.44 & 50.00 & 37.50 & 39.57 \\
    electrical engineer & 48.65 & 43.24 & 42.55 & 40.54 & 48.65 & 51.34 \\
    metrology engineer & 41.67 & 37.50 & 36.45 & 41.67 & 41.67 & 39.98 \\
    high school mathematics & 50.00 & 38.89 & 44.44 & 38.89 & 33.33 & 44.44 \\
    high school physics & 47.37 & 57.89 & 55.88 & 42.11 & 47.37 & 50.31 \\
    high school chemistry & 47.37 & 52.63 & 54.54 & 47.37 & 52.63 & 54.54 \\
    high school biology & 63.16 & 47.37 & 48.76 & 52.63 & 57.89 & 56.54 \\
    middle school mathematics & 63.16 & 47.37 & 65.76 & 63.16 & 42.11 & 65.76 \\
    middle school biology & 38.10 & 38.10 & 38.10 & 38.10 & 38.10 & 38.10 \\
    middle school physics & 36.84 & 57.89 & 59.87 & 31.58 & 47.37 & 48.76 \\
    middle school chemistry & 60.00 & 45.00 & 45.00 & 60.00 & 35.00 & 43.32 \\
    veterinary medicine & 52.17 & 43.48 & 51.11 & 47.83 & 39.13 & 51.78 \\
    college economics & 29.09 & 36.36 & 34.87 & 27.27 & 29.09 & 31.23 \\
    business administration & 27.27 & 36.36 & 34.44 & 30.30 & 36.36 & 41.21 \\
    marxism & 42.10 & 52.63 & 54.67 & 42.11 & 57.89 & 57.43 \\
    mao zedong thought & 29.17 & 25.00 & 25.00 & 33.33 & 25.00 & 24.43 \\
    education science & 31.03 & 20.69 & 33.43 & 31.03 & 27.59 & 34.32 \\
    teacher qualification & 22.73 & 27.27 & 29.98 & 27.27 & 31.82 & 32.77 \\
    high school politics & 47.37 & 52.63 & 65.47 & 63.16 & 57.89 & 68.98 \\
    high school geography & 47.37 & 52.63 & 53.21 & 42.11 & 52.63 & 53.21 \\
    middle school politics & 38.10 & 38.10 & 38.10 & 38.10 & 42.86 & 37.81 \\
    middle school geography & 50.00 & 75.00 & 71.10 & 50.00 & 58.33 & 45.78 \\
    modern chinese history & 34.78 & 47.83 & 46.43 & 39.13 & 56.52 & 62.12 \\
    ideological and moral cultivation & 57.89 & 42.11 & 45.55 & 52.63 & 31.58 & 42.21 \\
    logic & 54.55 & 40.91 & 54.55 & 54.55 & 36.36 & 54.55 \\
    law & 37.50 & 45.83 & 47.76 & 45.83 & 54.17 & 52.32 \\
    chinese language and literature & 43.48 & 56.52 & 55.55 & 43.48 & 56.52 & 55.55 \\
    art studies & 39.39 & 27.27 & 31.21 & 30.30 & 30.30 & 34.44 \\
    professional tour guide & 41.38 & 51.72 & 49.87 & 37.93 & 41.38 & 45.78 \\
    legal professional & 47.83 & 52.17 & 53.43 & 47.83 & 56.52 & 56.66 \\
    high school chinese & 42.10 & 47.37 & 45.56 & 36.84 & 47.37 & 48.64 \\
    high school history & 50.00 & 30.00 & 45.00 & 50.00 & 45.00 & 55.00 \\
    middle school history & 40.91 & 50.00 & 49.94 & 40.91 & 36.36 & 45.77 \\
    civil servant & 25.53 & 31.91 & 30.76 & 25.53 & 36.17 & 34.77 \\
    sports science & 47.37 & 42.11 & 45.54 & 57.89 & 63.16 & 70.12 \\
    plant protection & 40.91 & 50.00 & 45.88 & 45.45 & 45.45 & 45.88 \\
    basic medicine & 57.89 & 63.16 & 65.67 & 52.63 & 57.89 & 61.32 \\
    clinical medicine & 50.00 & 50.00 & 50.00 & 40.91 & 40.91 & 40.91 \\
    urban and rural planner & 41.30 & 45.65 & 41.30 & 41.30 & 32.61 & 41.30 \\
    accountant & 28.57 & 36.73 & 32.44 & 22.45 & 38.78 & 34.34 \\
    fire engineer & 35.48 & 32.26 & 30.32 & 29.03 & 45.16 & 37.75 \\
    environmental impact assessment & 45.16 & 41.94 & 46.56 & 45.16 & 32.26 & 46.56 \\
    tax accountant & 34.69 & 30.61 & 34.69 & 32.65 & 32.65 & 32.65 \\
    physician & 22.45 & 34.69 & 35.32 & 28.57 & 38.78 & 38.78 \\
    Average & 42.96 & 44.10 & \textbf{45.88} & 42.26 & 43.73 & \textbf{46.86} \\
    \bottomrule
    \end{tabular}
\end{table}
\clearpage

\subsection{Associative Recall}
\label{sec:associative_recall}

In this section, we trained a 1B parameter Jamba model and a 1B parameter OTCE model on 80\% of the pre-training data set with a sequence length of 4096, and then continued to pre-train the remaining 20\% of the pre-training data set as a 128K context length data set to continue training the model. The fine-tuning data set is also packaged as a 128K context length data set to fine-tune the model. The two models are extended to 128K context length for evaluation. The parameter settings of the two models are shown in Table~\ref{tab:associative_recall_params}.

\begin{table}[!ht]
    \footnotesize
    \centering
    \caption{
        (\textbf{Model Parameters for Associative Recall}.)
        The specific implementation of Jamba~\citep{lieber2024jamba} is exactly the same as the original paper, without using the improved modules in this paper. 
        Also, because the proportion of mixed experts is different, we adjust the dimension of $FFN$ in Jamba so that the total number of parameters is the same as OTCE.
    }
    \label{tab:associative_recall_params}
    \begin{tabularx}{0.9\linewidth}{@{}ccccccccc@{}}
    \toprule
    \sc{Model} & \sc{Layers} & \sc{Model Dim} & \sc{Attn Heads} & \sc{Group} & \sc{State Dim} & \sc{Experts} & \sc{Experts topK} & \sc{Experts Type} \\
    \midrule
    Jamba & 24 & 1024 & 16 & 2 & 64 & 4 & 1 & - \\
    OTCE & 24 & 1024 & 16 & 2 & 64 & 4 & 1 & Expansive \\
    \bottomrule
    \end{tabularx}
    
    \vspace{0em}

    \begin{tabularx}{0.9\linewidth}{@{}cc@{}}
    \toprule
    \sc{Model} & \sc{Hybrid} \\
    & \sc{s=ssm a=attention m=mlp e=moe} \\
    \midrule
    Jamba & $(\SM + \SE + \SM + \SE + \AM + \SE + \SM + \SE) \times 3$ \\
    OTCE & $\SM + \SE \times 11 + \AE \times 2 + \SE \times 8 + \SM + \AM$ \\
    \bottomrule
    \end{tabularx}
\end{table}

\begin{table}[!ht]
    \centering
    \caption{
        (\textbf{Evaluations for Associative Recall}.)
        We select the evaluations for associative recall including: 
        standard short context task: PIQA~\citep{bisk2020piqa},
        natural long context task: NarrativeQA~\citep{kovcisky2018narrativeqa},
        synthetic long context task: HotpotQA~\citep{yang2018hotpotqa},
        to evaluate the performance of Jamba and OTCE on these three tasks.
    }
    \label{tab:associative_recall_evaluations}
    \begin{tabular}{@{}lllllll@{}}
    \toprule
    \sc{Model} & \sc{PIQA} & \sc{NarrativeQA} & \sc{HotpotQA} & \sc{Average} \\
    & \sc{acc $\uparrow$} & \sc{acc $\uparrow$} & \sc{acc $\uparrow$} & \sc{acc $\uparrow$} \\
    \midrule
    Jamba & 78.65 & 23.15 & 34.40 & 45.4 \\
    OTCE & 79.48 & 27.98 & 42.71 & 50.0 \\
    \bottomrule
    \end{tabular}
\end{table}

In Table~\ref{tab:associative_recall_evaluations}, we show the performance of Jamba and OTCE on standard short context tasks, natural long context tasks, and synthetic long context tasks. It is known that Jamba~\citep{lieber2024jamba} can be successfully trained on tasks with a context length of 1M, not only performing well on tasks with a context length of 256K, but also surpassing many well-known open-source models such as Llama~\citep{touvron2023llama2}, Gemma~\citep{team2024gemma}, Mixtral~\citep{jiang2024mixtral} in common tasks. Therefore, we believe that Jamba is an excellent baseline model, and comparing with it is equivalent to comparing with many well-known open-source models.

In standard short context tasks, it is obvious that the model does not need to have the ability to handle long-distance dependencies. Our goal is to evaluate whether the accuracy of short-distance associative recall will be affected when the OTCE model is extended to handle long context environments in common tasks. The experimental results show that OTCE outperforms Jamba in standard short context tasks. Therefore, while adapting to long context environments, OTCE does not sacrifice accuracy in common tasks.

In natural long context tasks, OTCE achieved a performance improvement of up to $20.86\%$ compared to Jamba. We analyze that the main reason for this improvement is that SSM has a learnable matrix $(\dt, B, C)$ during training. However, in the inference stage, although these matrices can be dynamically calculated according to different input data, the parameters used for these calculations (i.e., the function or mapping that determines how to calculate these matrices and step sizes) are fixed, and these parameters are reused in the inference stage after being determined in the training stage. This means that the state of SSM may still be affected by irrelevant documents in the inference stage. In contrast, the last layer of Jamba consists of SSM and MOE. In the inference stage, not only the output state may be confused when approaching the output, but also as the ablation experiment of MOE and MLP in the previous section~\ref{tab:MOE_vs_MLP_Results} shows, the output state may also be unstable. These factors together lead to Jamba's performance in handling long context tasks is not as good as OTCE.

We all know that the Transformer model tends to copy the answers in the context examples rather than predict the answers to the actual questions. In contrast, the SSM model tends to predict answers rather than directly copy answers from the context. In synthetic long context tasks, the model needs to recall information from lengthy irrelevant text, requiring the ability to track across contexts and aggregate information. The reason why OTCE outperforms Jamba by $27.06\%$ in synthetic long context tasks is similar to the attention decay and knowledge routing bias issues discussed in the previous section~\ref{sec:attention_before_output_is_important}. The Expresser module of OTCE reweights the recursively aggregated information with gradually decaying long-term dependency relationships through re-attention to all elements, and learns how SSM recursively aggregates information to correct knowledge routing bias. This allows OTCE to outperform Jamba in synthetic long context tasks.

\begin{figure}[!ht]
    \centering
    \caption{
        (\textbf{Needle in a haystack}.)
        OTCE can recall information across 128K context length (our maximum extended sequence length), and has excellent performance on the Needle in a haystack evaluation.
    }
    \includegraphics[width=1.0\textwidth]{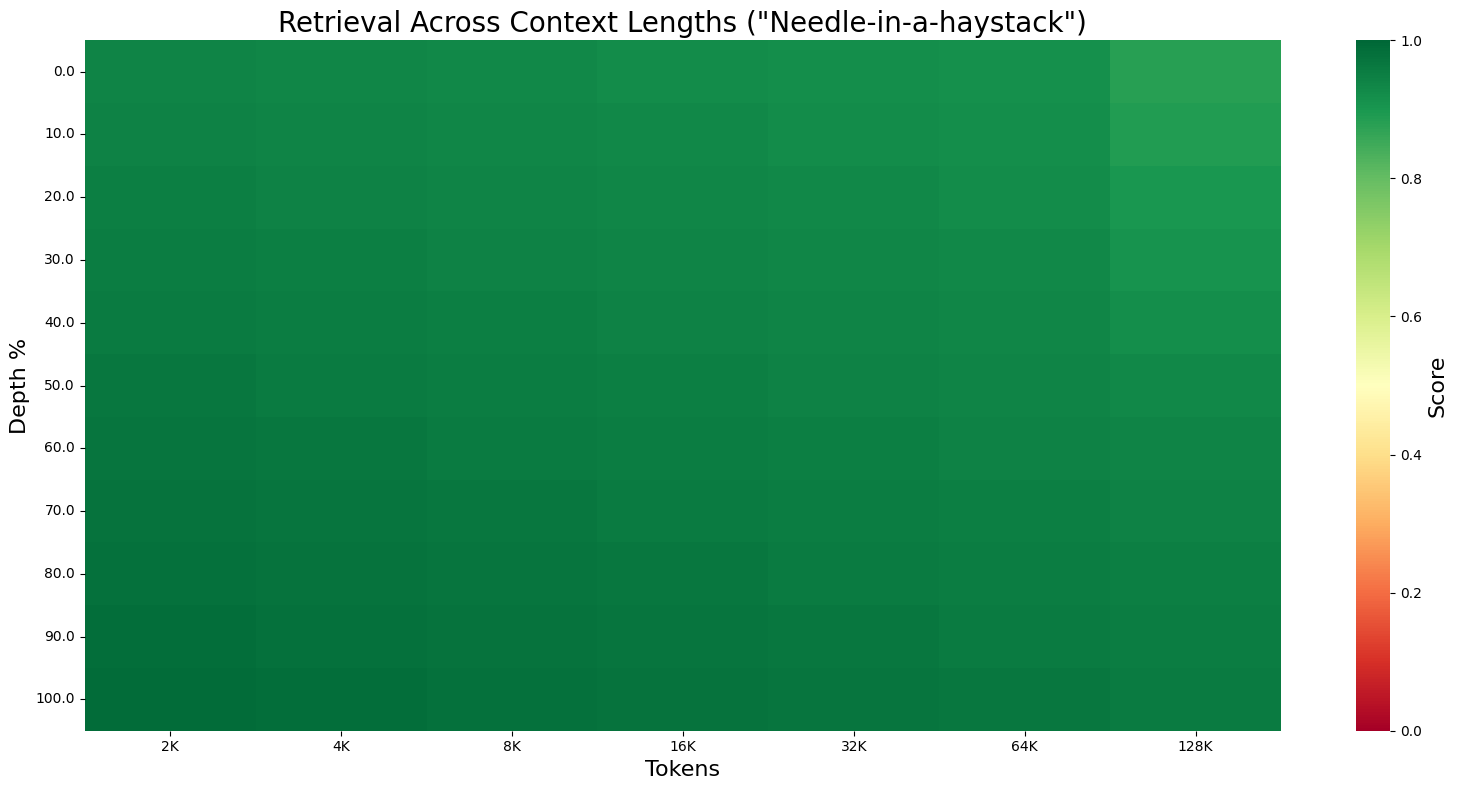}
    \label{fig:needle}
\end{figure}

In the Needle in a haystack problem, the model is required to recall information from lengthy irrelevant text input, as shown in Figure~\ref{fig:needle}. OTCE has excellent performance on the Needle in a haystack evaluation.

  \section{Conclusion}
\label{sec:conclusion}

We propose a position information encoding method that connects SSM and Attention, a sparse cross-domain mixture of experts with higher learning efficiency, and design a hybrid SSM and Attention sparse model architecture: OTCE with a biomimetic idea, achieving state-of-the-art performance in language modeling.

  \printbibliography

  \newpage

  \appendix

  \onecolumn
  
  \section{RoPE for SSM}
\label{sec:rope_for_ssm}

\begin{proof}[Proof of \cref{eq:ssm_rope_g_final}]
The rotation position encoding for $C$ and $B$ matrices in SSM.

\begin{subequations}
    \begin{align*}
        C_{j} &= f_{C} (x_{j}, j) \\
        B_{i} &= f_{B} (x_{i}, i)
    \end{align*}
\end{subequations}

$C_{j}$ represents the output weight matrix of the $j$-th token corresponding to the word vector $x_{j}$ integrated with the position information $j$,
$B_{i}$ represents the input weight matrix of the $i$-th token corresponding to the word vector $x_{i}$ integrated with the position information $i$.

To utilize the relative positional information between tokens,
we assume that the inner product operation between the $C_{j}$ vector and the $B_{i}$ vector can be represented by a function $g$,
where the input of the function $g$ is the word embedding vectors $x_{j}$ and $x_{i}$,
and their relative positional information $j - i$,
the inner product of $C_{j}$ and $B_{i}$ and their relative positional information $j - i$ is defined as

\begin{equation*}
    <f_{C}(x_{j}, j), f_{B}(x_{i}, i)> = g(x_{j}, x_{i}, j - i)
\end{equation*}

Now, assuming the word embedding vector dimension is $d = 2$,
we have $f_{C}(x_{j}, j) = (W_{C} x_{j})e^{\imath j \theta}$,
for the first half of the formula $W_{C} x_{j}$,
we know that $W_{C}$ is a two-dimensional matrix,
$x_{j}$ is a two-dimensional vector,
the result of the multiplication is naturally a two-dimensional vector,
represented by $C_{j}$

\begin{align*}
    C_{j} &= 
        \begin{bmatrix} 
            C_{j}^{(1)} \\
            C_{j}^{(2)}
        \end{bmatrix}
    = W_{C} x_{j} = 
        \begin{bmatrix} 
            W_{C}^{(11)} & W_{C}^{(12)} \\
            W_{C}^{(21)} & W_{C}^{(22)}
        \end{bmatrix}
        \begin{bmatrix} 
            x_{j}^{(1)} \\
            x_{j}^{(2)}
        \end{bmatrix}
\end{align*}

For the second half $e^{\imath j \theta}$,
according to Euler's formula $e^{\imath x} = \cos(x) + \imath \sin(x)$,
we have

\begin{align*}
    e^{\imath j \theta} &= \cos(j \theta) + \imath \sin(j \theta)
\end{align*}

We know

\begin{align*}
    f_{C}(x_{j}, j) &= (W_{C} x_{j})e^{\imath j \theta} = C_{j}e^{\imath j \theta}
\end{align*}

$C_{j}$ is represented in complex form,

\begin{align*}
    C_{j} &=
        \begin{bmatrix} 
            C_{j}^{(1)}, C_{j}^{(2)}
        \end{bmatrix}
    = 
        \begin{bmatrix} 
            C_{j}^{(1)} + \imath C_{j}^{(2)}
        \end{bmatrix}
\end{align*}

Thus,

\begin{align*}
    f_{C}(x_{j}, j) &= C_{j}e^{\imath j \theta} = 
        \begin{bmatrix} 
            C_{j}^{(1)} + \imath C_{j}^{(2)}
        \end{bmatrix} e^{\imath j \theta}
\end{align*}

According to the above derivation,
we know that $f_{C}(x_{j}, j)$ is the product of two complex numbers,

\begin{align*}
    f_{C}(x_{j}, j) &= C_{j}e^{\imath j \theta} = 
        \begin{bmatrix} 
            C_{j}^{(1)} + \imath C_{j}^{(2)}
        \end{bmatrix} \times (\cos(j \theta) + \imath \sin(j \theta))
\end{align*}

Considering the following two formulas about complex numbers

\begin{align*}
    (a + \imath b) \times (c + \imath d) &= ac + \imath bc + \imath ad + \imath^2 bd = (ac - bd) + \imath (bc + ad) \\
    \imath^2 &= -1
\end{align*}

We have

\begin{align*}
    C_{j}e^{\imath j \theta} &= 
        \begin{bmatrix} 
            C_{j}^{(1)} + \imath C_{j}^{(2)}
        \end{bmatrix} \times (\cos(j \theta) + \imath \sin(j \theta)) = 
        \begin{bmatrix} 
            C_{j}^{(1)} \cos(j \theta) - C_{j}^{(2)} \sin(j \theta)
        \end{bmatrix} + \imath
        \begin{bmatrix} 
            C_{j}^{(2)} \cos(j \theta) + C_{j}^{(1)} \sin(j \theta)
        \end{bmatrix}
\end{align*}

Expressing this result as a real vector,

\begin{align*}
    C_{j}e^{\imath j \theta} &= 
        \begin{bmatrix} 
            C_{j}^{(1)} \cos(j \theta) - C_{j}^{(2)} \sin(j \theta),
            C_{j}^{(2)} \cos(j \theta) + C_{j}^{(1)} \sin(j \theta)
        \end{bmatrix}
\end{align*}

Therefore, $C_{j}$ multiplied by a rotation matrix is obtained.

\begin{align*}
    f_{C}(x_{j}, j) &= (W_{C} x_{j})e^{\imath j \theta} = C_{j}e^{\imath j \theta} \\
    &=
    \begin{bmatrix} 
        C_{j}^{(1)} \cos(j \theta) - C_{j}^{(2)} \sin(j \theta),
        C_{j}^{(2)} \cos(j \theta) + C_{j}^{(1)} \sin(j \theta)
    \end{bmatrix} \\
    &=
    \begin{bmatrix} 
        \cos(j \theta) & -\sin(j \theta) \\
        \sin(j \theta) & \cos(j \theta)
    \end{bmatrix}
    \begin{bmatrix} 
        C_{j}^{(1)} \\
        C_{j}^{(2)}
    \end{bmatrix}
\end{align*}

Similarly, $B_{i}$ vector can be obtained

\begin{align*}
    f_{B}(x_{i}, i) &= (W_{B} x_{i})e^{\imath i \theta} = B_{i}e^{\imath i \theta} \\
    &=
    \begin{bmatrix} 
        B_{i}^{(1)} \cos(i \theta) - B_{i}^{(2)} \sin(i \theta),
        B_{i}^{(2)} \cos(i \theta) + B_{i}^{(1)} \sin(i \theta)
    \end{bmatrix} \\
    &=
    \begin{bmatrix} 
        \cos(i \theta) & -\sin(i \theta) \\
        \sin(i \theta) & \cos(i \theta)
    \end{bmatrix}
    \begin{bmatrix} 
        B_{i}^{(1)} \\
        B_{i}^{(2)}
    \end{bmatrix}
\end{align*}

The function $g$ can be represented as

\begin{align*}
    g(x_{j}, x_{i}, j - i) &= \Re 
        \begin{bmatrix} 
            (W_{C} x_{j})(W_{B} x_{i})^*e^{\imath (j - i) \theta}
        \end{bmatrix}
\end{align*}

where $\Re$ represents the real part of the complex number $x$,
$(W_{C} x_{j})(W_{B} x_{i})^*$ represents the conjugate of the product of two complex numbers.
Considering

\begin{align*}
    z &= a + \imath b \\
    z^* &= a - \imath b
\end{align*}

we have

\begin{align*}
    W_{C} x_{j} &= C_{j} = C_{j}^{(1)} + \imath C_{j}^{(2)} \\
    W_{B} x_{i} &= B_{i} = B_{i}^{(1)} + \imath B_{i}^{(2)} \\
    (W_{B} x_{i})^* &= B_{i}^* = B_{i}^{(1)} - \imath B_{i}^{(2)} \\
    e^{\imath (j - i) \theta} &= \cos((j - i) \theta) + \imath \sin((j - i) \theta)
\end{align*}

We now want to prove that

\begin{align*}
    g(x_{j}, x_{i}, j - i) &= \Re 
        \begin{bmatrix} 
            (W_{C} x_{j})(W_{B} x_{i})^*e^{\imath (j - i) \theta}
        \end{bmatrix} \\
    &= \Re
        \begin{bmatrix} 
            (C_{j}^{(1)} + \imath C_{j}^{(2)})(B_{i}^{(1)} - \imath B_{i}^{(2)})(\cos((j - i) \theta) + \imath \sin((j - i) \theta))
        \end{bmatrix} \\
    &= \Re
        \begin{bmatrix} 
            ((C_{j}^{(1)}B_{i}^{(1)} + C_{j}^{(2)}B_{i}^{(2)}) + \imath (C_{j}^{(2)}B_{i}^{(1)} - C_{j}^{(1)}B_{i}^{(2)}))(\cos((j - i) \theta) + \imath \sin((j - i) \theta))
        \end{bmatrix} \\
    &= (C_{j}^{(1)}B_{i}^{(1)} + C_{j}^{(2)}B_{i}^{(2)})\cos((j - i) \theta) - (C_{j}^{(2)}B_{i}^{(1)} - C_{j}^{(1)}B_{i}^{(2)})\sin((j - i) \theta)
\end{align*}

Recalling the vectorized form of SSM~\ref{eq:ssm-matrix-vectorized},
the $C$ vector at position $j$ and the $B$ vector at position $i$ will perform an inner product operation,
that is,

\begin{align*}
    f_{C}(x_{j}, j) &= 
        \begin{bmatrix} 
            C_{j}^{(1)} \cos(j \theta) - C_{j}^{(2)} \sin(j \theta),
            C_{j}^{(2)} \cos(j \theta) + C_{j}^{(1)} \sin(j \theta)
        \end{bmatrix} \\
    f_{B}(x_{i}, i) &=
        \begin{bmatrix} 
            B_{i}^{(1)} \cos(i \theta) - B_{i}^{(2)} \sin(i \theta),
            B_{i}^{(2)} \cos(i \theta) + B_{i}^{(1)} \sin(i \theta)
        \end{bmatrix}
\end{align*}

We have

\begin{align*}
    <f_{C}(x_{j}, j), f_{B}(x_{i}, i)> &= 
        \begin{bmatrix} 
            C_{j}^{(1)} \cos(j \theta) - C_{j}^{(2)} \sin(j \theta)
        \end{bmatrix}
        \begin{bmatrix} 
            B_{i}^{(1)} \cos(i \theta) - B_{i}^{(2)} \sin(i \theta)
        \end{bmatrix} \\
        &+
        \begin{bmatrix} 
            C_{j}^{(2)} \cos(j \theta) + C_{j}^{(1)} \sin(j \theta)
        \end{bmatrix}
        \begin{bmatrix} 
            B_{i}^{(2)} \cos(i \theta) + B_{i}^{(1)} \sin(i \theta)
        \end{bmatrix} \\
        &= C_{j}^{(1)} \cos(j \theta) B_{i}^{(1)} \cos(i \theta) - C_{j}^{(1)} \cos(j \theta) B_{i}^{(2)} \sin(i \theta) \\
        &- C_{j}^{(2)} \sin(j \theta) B_{i}^{(1)} \cos(i \theta) + C_{j}^{(2)} \sin(j \theta) B_{i}^{(2)} \sin(i \theta) \\
        &+ C_{j}^{(2)} \cos(j \theta) B_{i}^{(2)} \cos(i \theta) + C_{j}^{(2)} \cos(j \theta) B_{i}^{(1)} \sin(i \theta) \\
        &+ C_{j}^{(1)} \sin(j \theta) B_{i}^{(2)} \cos(i \theta) + C_{j}^{(1)} \sin(j \theta) B_{i}^{(1)} \sin(i \theta)
\end{align*}

Considering

\begin{align*}
    \sin(a + b) &= \sin(a)\cos(b) + \cos(a)\sin(b) \\
    \sin(a - b) &= \sin(a)\cos(b) - \cos(a)\sin(b) \\
    \cos(a + b) &= \cos(a)\cos(b) - \sin(a)\sin(b) \\
    \cos(a - b) &= \cos(a)\cos(b) + \sin(a)\sin(b)
\end{align*}

We have

\begin{align*}
    <f_{C}(x_{j}, j), f_{B}(x_{i}, i)> &= 
        C_{j}^{(1)} B_{i}^{(1)} (\cos(j \theta) \cos(i \theta) + \sin(j \theta) \sin(i \theta)) \\
        &+ C_{j}^{(1)} B_{i}^{(2)} (-\cos(j \theta) \sin(i \theta) + \sin(j \theta) \cos(i \theta)) \\
        &+ C_{j}^{(2)} B_{i}^{(1)} (-\sin(j \theta) \cos(i \theta) + \cos(j \theta) \sin(i \theta)) \\
        &+ C_{j}^{(2)} B_{i}^{(2)} (\sin(j \theta) \sin(i \theta) + \cos(j \theta) \cos(i \theta)) \\
        &= C_{j}^{(1)} B_{i}^{(1)} \cos((j - i) \theta) + C_{j}^{(1)} B_{i}^{(2)} \sin((j - i) \theta) \\
        &- C_{j}^{(2)} B_{i}^{(1)} \sin((j - i) \theta) + C_{j}^{(2)} B_{i}^{(2)} \cos((j - i) \theta) \\
        &= (C_{j}^{(1)} B_{i}^{(1)} + C_{j}^{(2)} B_{i}^{(2)})\cos((j - i) \theta) + (C_{j}^{(1)} B_{i}^{(2)} - C_{j}^{(2)} B_{i}^{(1)})\sin((j - i) \theta) \\
        &= (C_{j}^{(1)}B_{i}^{(1)} + C_{j}^{(2)}B_{i}^{(2)})\cos((j - i) \theta) - (C_{j}^{(2)}B_{i}^{(1)} - C_{j}^{(1)}B_{i}^{(2)})\sin((j - i) \theta) \\
        &= g(x_{j}, x_{i}, j - i)
\end{align*}

It is proved that the inner product of the $C$ vector at position $j$ and the $B$ vector at position $i$ is the function $g$.

Finally, using the matrix-vector multiplication form

\begin{align*}
    <f_{C}(x_{j}, j), f_{B}(x_{i}, i)> &= 
        \begin{bmatrix} 
            \begin{bmatrix} 
                \cos(j \theta) & -\sin(j \theta) \\
                \sin(j \theta) & \cos(j \theta)
            \end{bmatrix}
            \begin{bmatrix} 
                C_{j}^{(1)} \\
                C_{j}^{(2)}
            \end{bmatrix}
        \end{bmatrix}^{T}
        \begin{bmatrix} 
            \begin{bmatrix} 
                \cos(i \theta) & -\sin(i \theta) \\
                \sin(i \theta) & \cos(i \theta)
            \end{bmatrix}
            \begin{bmatrix} 
                B_{i}^{(1)} \\
                B_{i}^{(2)}
            \end{bmatrix}
        \end{bmatrix} \\
        &= 
        \begin{bmatrix} 
            C_{j}^{(1)} & C_{j}^{(2)}
        \end{bmatrix}
        \begin{bmatrix} 
            \cos(j \theta) & \sin(j \theta) \\
            -\sin(j \theta) & \cos(j \theta)
        \end{bmatrix}
        \begin{bmatrix} 
            \cos(i \theta) & -\sin(i \theta) \\
            \sin(i \theta) & \cos(i \theta)
        \end{bmatrix}
        \begin{bmatrix} 
            B_{i}^{(1)} \\
            B_{i}^{(2)}
        \end{bmatrix} \\
\end{align*}

Expanding the product of the two rotation matrices, we have

\begin{align*}
    \begin{bmatrix} 
        \cos(j \theta) \cos(i \theta) + \sin(j \theta) \sin(i \theta) & -\cos(j \theta) \sin(i \theta) + \sin(j \theta) \cos(i \theta) \\
        -\sin(j \theta) \cos(i \theta) + \cos(j \theta) \sin(i \theta) & \sin(j \theta) \sin(i \theta) + \cos(j \theta) \cos(i \theta)
    \end{bmatrix}
\end{align*}

Finally, we get

\begin{align*}
    <f_{C}(x_{j}, j), f_{B}(x_{i}, i)> &= 
        \begin{bmatrix} 
            C_{j}^{(1)} & C_{j}^{(2)}
        \end{bmatrix}
        \begin{bmatrix} 
            \cos((j - i) \theta) & -\sin((j - i) \theta) \\
            \sin((j - i) \theta) & \cos((j - i) \theta)
        \end{bmatrix}
        \begin{bmatrix} 
            B_{i}^{(1)} \\
            B_{i}^{(2)}
        \end{bmatrix}
\end{align*}

The above derivation is only for the case of word embedding dimension $d = 2$,
when $d > 2$, the two-dimensional case can be extended to any dimension as follows

\begin{align*}
    f_{\{C, B\}}(x_{j}, j) &= \R_{\Theta, j}^{d} W_{\{C, B\}} x_{j}
\end{align*}

The inner product satisfies linearity,
so for any even-dimensional RoPE, we can represent it as a concatenation of the two-dimensional case,
that is, grouping the elements of the word embedding vector in pairs

\begin{align*}
    \R_{\Theta, j}^{d} = \begin{bmatrix} 
        \cos j \theta_0 & -sin j \theta_0 & 0 & 0 & \dots & 0 & 0 \\
        \sin j \theta_0 & \cos j \theta_0 & 0 & 0 & \dots & 0 & 0 \\
        0 & 0 & \cos j \theta_1 & -sin j \theta_1 & \dots & 0 & 0 \\
        0 & 0 & \sin j \theta_1 & \cos j \theta_1 & \dots & 0 & 0 \\
        \vdots & \vdots & \vdots & \vdots & \ddots & \vdots & \vdots \\
        0 & 0 & 0 & 0 & \dots & \cos j \theta_{d/2} & -sin j \theta_{d/2-1} \\
        0 & 0 & 0 & 0 & \dots & \sin j \theta_{d/2} & \cos j \theta_{d/2-1} \\
    \end{bmatrix}
\end{align*}

Each group applies the same rotation operation and the rotation angle of each group is calculated as follows:

\begin{align*}
    \Theta &= \{\theta_i = 10000^{-2(i - 1) / d}, i \in [1, 2, \dots, d / 2]\}
\end{align*}

RoPE is a kind of relative positional encoding, and relative positional encoding is a special form of Toeplitz matrix

\begin{align*}
    \begin{bmatrix} 
        0 & & & & & & \\
        1 & 0 & & & & \\
        2 & 1 & 0 & & \\
        3 & 2 & 1 & 0 \\
        \vdots & \ddots & \ddots & \ddots & \ddots \\
        n-1 & n-2 & n-3 & \dots & 1 & 0
    \end{bmatrix}
\end{align*}

We can know that the position distribution after RoPE is unbalanced, 0 appears most frequently as the low bit, and n-1 appears least frequently as the high bit, which also leads to a problem of RoPE, its high bit is not sufficiently trained, and the generalization ability is not as good as the low bit. We take the average value of the effective sequence length of the training data as the base, for the sequence length greater than the base, we have

\begin{align*}
    C \times \max(1, \log_{base} n) \\
\end{align*}

The part of the sequence length within the base is not affected, and the part of the sequence length greater than the base is expanded according to the ratio of $\log_{base} n$, so that the problem of insufficient generalization ability of the high bit can be solved.

\end{proof}
\end{document}